\documentclass{tADR2e}

\def\vector#1{\mbox{\boldmath $#1$}}
\DeclareMathOperator*{\argmax}{argmax}

\usepackage{enumerate}
\usepackage{algorithm,algorithmic}
\usepackage{comment}

\begin{document}

\jvol{00} \jnum{00} \jyear{2021} \jmonth{March}

%\articletype{GUIDE}

\title{Hierarchical Bayesian Model for the Transfer of Knowledge on Spatial Concepts based on Multimodal Information}

\author{Yoshinobu Hagiwara$^{a}$$^{\ast}$\thanks{$^\ast$Corresponding author. Email: yhagiwara@em.ci.ritsumei.ac.jp
\vspace{6pt}}, Keishiro Taguchi$^{a}$, Satoshi Ishibushi$^{b}$, Akira Taniguchi$^{a}$, and Tadahiro Taniguchi$^{a}$\\
\vspace{6pt}$^{a}${\em{Ritsumeikan University \\ 1-1-1 Noji Higashi, Kusatsu, Shiga 525-8577, Japan}};\\
$^{b}${\em{Panasonic Corporation, Osaka, Japan}}\\\vspace{6pt}\received{v1.0 released December 2020} }

\maketitle

%Abstracts of up to 100 words for Short Papers or 200 words for Full/Survey Papers are required for all papers submitted.
\begin{abstract}
This paper proposes a hierarchical Bayesian model based on spatial concepts that enables a robot to transfer the knowledge of places from experienced environments to a new environment. The transfer of knowledge based on spatial concepts is modeled as the calculation process of the posterior distribution based on the observations obtained in each environment with the parameters of spatial concepts generalized to environments as prior knowledge. We conducted experiments to evaluate the generalization performance of spatial knowledge for general places such as kitchens and the adaptive performance of spatial knowledge for unique places such as `Emma's room' in a new environment. In the experiments, the accuracies of the proposed method and conventional methods were compared in the prediction task of location names from an image and a position, and the prediction task of positions from a location name. The experimental results demonstrated that the proposed method has a higher prediction accuracy of location names and positions than the conventional method owing to the transfer of knowledge.

\medskip

\end{abstract}

\section{Introduction}
\label{sec:introduction}
For a robot that supports human life to perform tasks in a new home environment, it is required to have the knowledge of general places in advance and adaptively acquire the knowledge of places unique to each home environment. 
For example, even in a new home environment, the robot has to predict appropriate positions from linguistic information such as `kitchen,' and to predict appropriate location names from observed information such as images and positions based on the knowledge of places.
If the robot can transfer the knowledge of general places (e.g., kitchen and living room) in various home environments, it can learn the general places only by acquiring partially observed data such as vision and position in the new home environment.
In addition, the robot requires the ability to adaptively learn places based on linguistic instructions from users and observations obtained in the new home environment because some home-specific places have unique features and location names, such as `Emma's room' and `father's room.'

In the field of image recognition, several models have been proposed to estimate the image class of a place using convolutional neural networks (CNNs)~\citep{Zhou14,Zhou18,Qassim18,Ursic16}.
In studies of semantic mapping, methods of assigning a place vocabulary or class to an occupied grid map~\citep{Kostavels15,Sunderhauf16,Pal19} and methods using a topological map have been proposed~\citep{Rangel17,Hiller19,Balaska20}.
These studies often use supervised learning techniques, such as CNNs, and require large-scale labeled datasets to learn model parameters.
For general places, these methods enable a robot to learn places by preparing an existing dataset (e.g., Places365~\cite{places365}), but for home-specific places such as Emma's room, preparing large-scale datasets to acquire the knowledge of these places is difficult.
To acquire such knowledge for home-specific places, this study focused on the learning of places based on linguistic instructions instead of labels.

In the field of robot navigation, studies on visual and language navigation have been proposed to achieve room-to-room navigation based on linguistic information using a dataset composed of image sequences with 21,576 linguistic instructions on navigation routes~\cite{Anderson18, Krantz20}. These studies achieved robot navigation based on the linguistic information of routes using a sequence-to-sequence model. However, since the model requires a large-scale dataset for image sequences with linguistic instructions on navigation routes, acquiring knowledge of home-specific places through on-site learning in a new environment is difficult. This study aimed to achieve on-site learning of places with a small amount of linguistic information.

Spatial concept models have been proposed to enable a robot to learn the knowledge of places based on the user's linguistic instructions and the robot's observations through on-site learning~\cite{Taniguchi16,Hagiwara18,Katsumata19}.
The spatial concept models learn spatial categories from the user's linguistic instructions and the robot's observations (i.e., position, language, and vision) through unsupervised learning based on a probabilistic generative model and Bayesian inference in each home environment.
These models enable a robot to learn home-specific places through on-site learning in each home environment.
However, in these studies, when the robot moves to a new home environment, the robot cannot reuse the knowledge of general places, e.g., `kitchen' and `living room,' acquired in experienced environments and must begin learning the new home environment from scratch. 
This instruction process imposes a heavy burden on the user.
To solve this problem, this study focused on an approach to transfer the knowledge of places acquired in experienced environments to the learning of new environments.

\begin{figure}[tb]
\begin{center}
    \includegraphics[width=450pt]{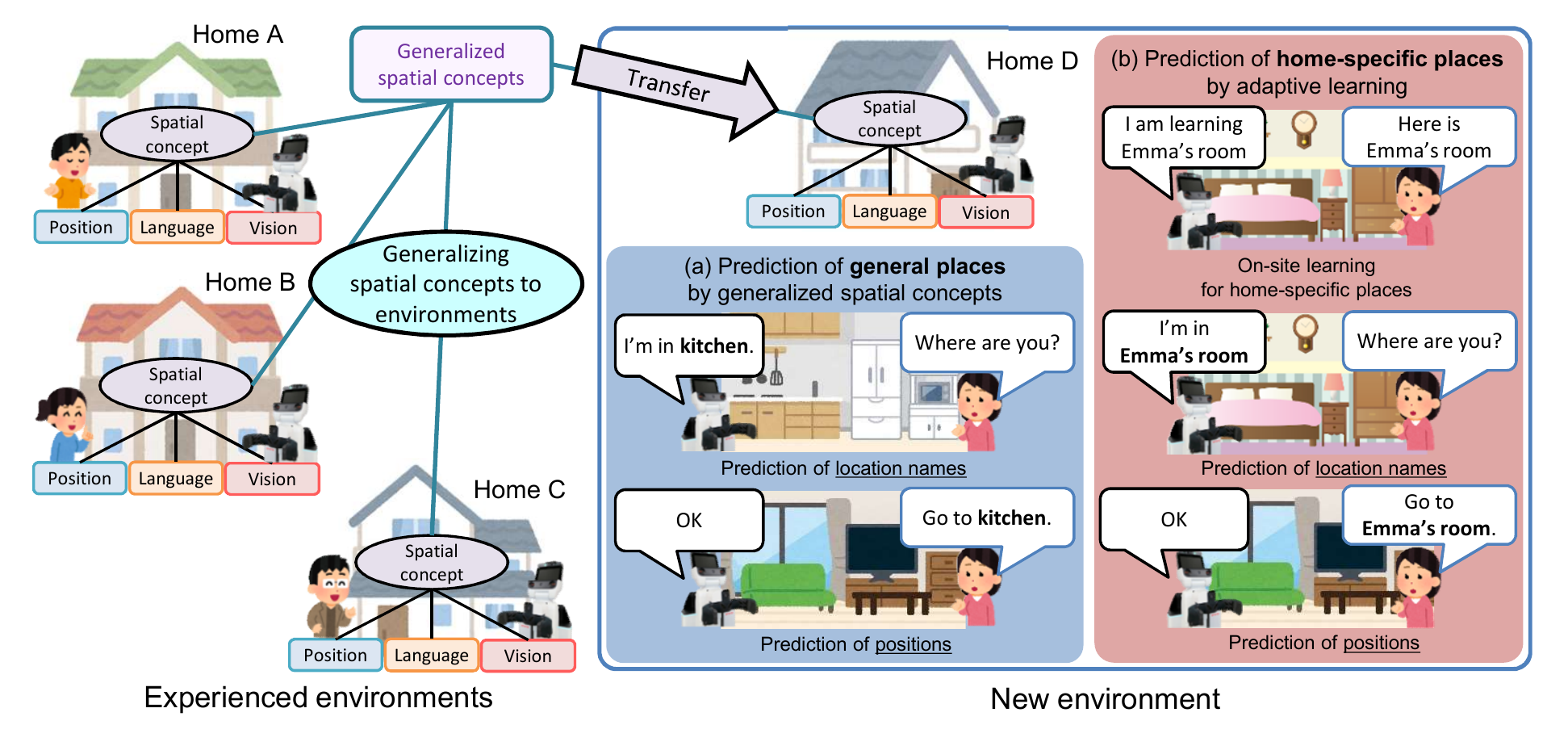}
    \caption{Transfer of the knowledge on spatial concepts from experienced environments to a new environment and the prediction of positions and names for general and home-specific places in the new environment.}
   \label{transfer}
  \end{center}
\end{figure}

In this paper, we propose a hierarchical Bayesian model based on spatial concepts that enables a robot to transfer the knowledge of places between environments and demonstrate the performance of name and position prediction using the proposed method in a new environment.
Fig.~\ref{transfer} shows an overview of the proposed method that transfers the knowledge on spatial concepts from experienced environments to a new one.
As the left side of Fig.~\ref{transfer} shows, the robot learns places as spatial concepts based on position, language, and vision information, and generalizes these spatial concepts.
In Fig.~\ref{transfer} (a), the robot learns spatial concepts for general places such as kitchens in a new home environment based on transferred spatial concepts and partially observed data (i.e., vision and position) without linguistic instructions from the user.
As Fig.~\ref{transfer} (b) shows, the proposed method also enables the robot to adaptively learn home-specific places (e.g., Emma's room) based on linguistic instructions from a user and its observations in the new environment. 
The prediction performance of the proposed method for general places and home-specific places was clarified from the accuracy of name and position predictions in experiments performed in various virtual home environments.

The main contributions of this paper are as follows:
\begin{itemize}
 \item We present a novel spatial concept model based on hierarchical Bayes modelling that transfers the knowledge of places obtained in experienced environments to the learning of places in a new environment.
 \item We clarify the performance of the proposed method in predicting general places from the accuracy of name and position predictions through a comparison with conventional spatial concept methods.
 \item We clarify the ability of the proposed method to adaptively learn home-specific places from the accuracy of name and position predictions in a step-wise change in the number of linguistic instructions.
\end{itemize}

The remainder of this paper is organized as follows. In Section 2, we present related studies on the acquisition of the knowledge of places. In Section 3, we present our method for transferring this knowledge between environments. In Section 4, we demonstrate an experiment to evaluate the accuracy of position and name predictions for general places based on transferred spatial concepts. In Section 5, we demonstrate an experiment to evaluate the accuracy of position and name predictions for home-specific places in the adaptive learning of spatial concepts. Finally, Section 6 concludes the paper. 

Details of related studies and its relationships with this study are provided in Appendix~\ref{apdx:related_studies}.

\section{Proposed method}

\subsection{Overview}
An overview of the transfer of knowledge of spatial concepts is shown in Fig.~\ref{transfer}. As shown on the left side of Fig.~\ref{transfer}, with the proposed method, the parameters of spatial concepts are inferred in each home environment based on observations (i.e., position and vision information) and linguistic instructions from the user (i.e., language information) as the knowledge from the source domain. As shown on the right-hand side of Fig.~\ref{transfer}, the parameters of spatial concepts related to language and vision information are generalized from experienced home environments to transfer knowledge on spatial concepts to the learning of places in a new home environment as a target domain. By using the knowledge on spatial concepts transferred from the experienced environments, the proposed method enables a robot to predict the location names of general places, such as `kitchen,' from observations partially obtained in the new home environment without linguistic instructions from the user (Fig.~\ref{transfer} (a)). Additionally, the proposed method enables the robot to adaptively learn home-specific places and regions, such as Emma's room, based on observations and linguistic instructions from the user in the new environment (Fig.~\ref{transfer} (b)). The performance of the proposed method was demonstrated through experiments to evaluate the accuracy of location name and position prediction in a new environment. 
In the proposed method, the robot formed spatial concepts based on multimodal information (i.e., vision and language) in each environment. An overview of the spatial concept formation based on multimodal information is provided  in Appendix~\ref{apdx:proposed_method:multimodal}.

\subsection{Model to transfer knowledge in the spatial concept model}
\begin{figure}[tb]
  \begin{center}
    \includegraphics[width=350pt]{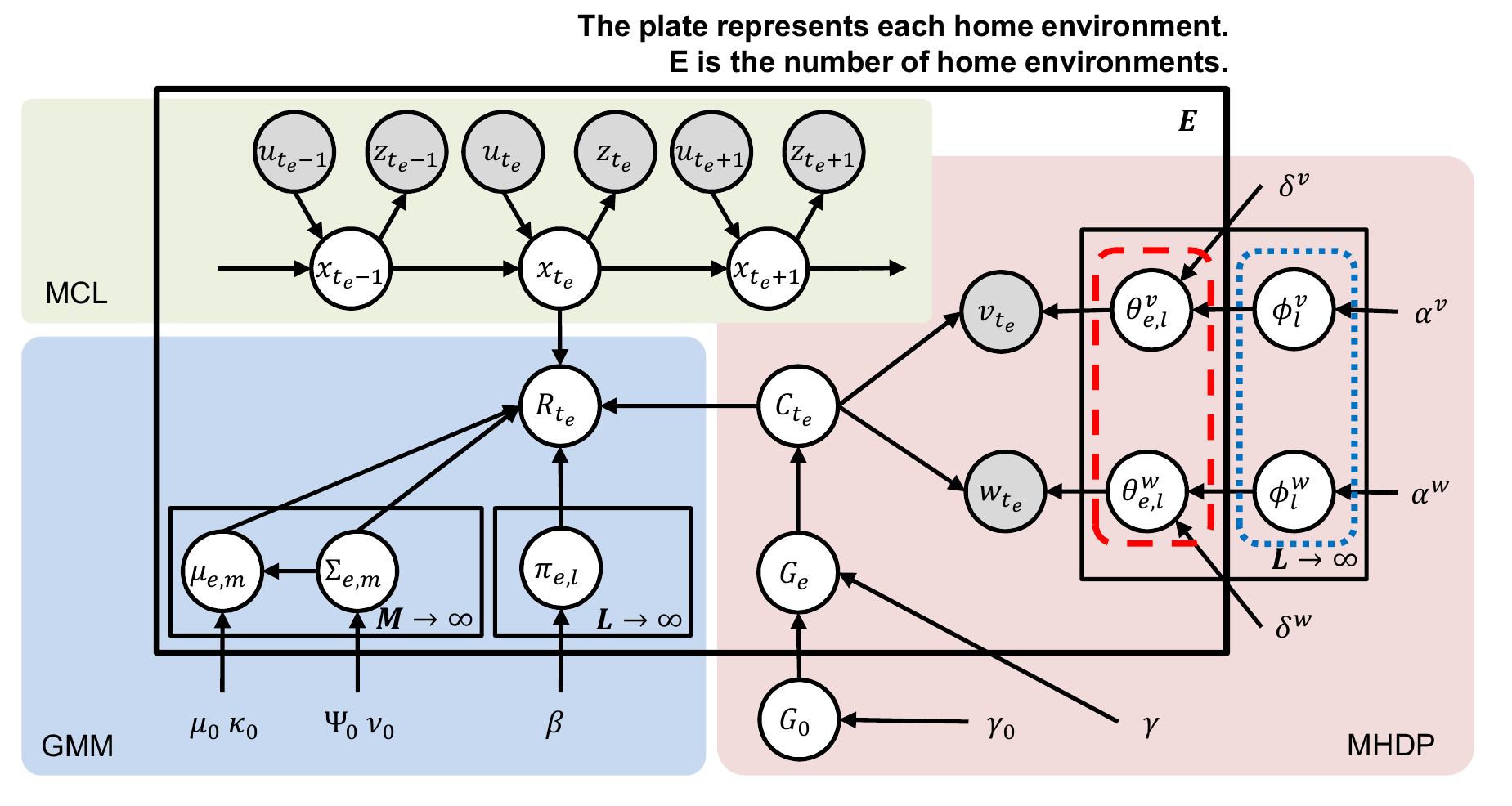}
    \caption{Graphical model of the proposed model extended for transferring knowledge between environments. The proposed model consists of an MCL to estimate a position ($x_{t_e}$), a GMM to estimate a spatial region ($R_{t_e}$), and an MHDP to estimate a spatial category ($C_{t_e}$) based on the knowledge transferred between the plates of environment $E$. $\phi^v_l$ and $\phi^w_l$ in the blue dotted line are the parameters inferred from the plates of $E$ as generalized knowledge. $\theta^v_{e,l}$ and $\theta^w_{e,l}$ in the red dashed line are parameters inferred in the plate of a specified environment $e$ based on prior distributions $\phi^v_l$ and $\phi^w_l$ as environment-specific knowledge.}
   \label{multi_model}
  \end{center}
\end{figure}

Fig.~\ref{multi_model} shows the graphical model of the proposed model consists of an MCL to estimate a position ($x_{t_e}$), a Gaussian mixture model (GMM) to estimate a spatial region ($R_{t_e}$), and a multimodal hierarchical Dirichlet process (MHDP) to estimate a spatial category ($C_{t_e}$) based on the knowledge transferred between environments.
The definitions of variables on the graphical model are shown in Table~\ref{model_val3}. 
$L$ and $M$ denote the number of spatial concepts and spatial regions, respectively, and $E$ denotes the number of environments in which a robot acquires observation. The number of spatial concepts and regions were estimated using the nonparametric Bayes model in the proposed model.

Since the SpCoA in Fig.~\ref{conv_model} does not have the environment plate ($E$), it is not possible to transfer the parameters of $R_t$ and $C_t$ from experienced environments to a new environment. In the model proposed in Fig.~\ref{multi_model}, the parameters $\theta$, $\mu$, and $\sigma$ that generate visual information $v_t$, linguistic information $w_t$, and a region $R_t$ are placed in $E$, and the parameters of the environment-specific $R_t$ and $C_t$ can be learned. Furthermore, by arranging the parameters $\phi$ that generate visual and linguistic information based on the spatial concept outside $E$, the parameters of spatial concepts generalized to environments are inferred and transferred as prior knowledge to the learning of places in a new environment.

\begin{table}[tb]
\caption{Description of the random variables used in the graphical model}
\begin{center}
\scalebox{0.8}{
\begin{tabular}{c p{15cm}}
\toprule
Symbol & Definition\\
\colrule
$u_{t_e}$& Control information\\
$z_{t_e}$& Measured depth information\\
$x_{t_e}$& Positional information of a robot\\
$v_{t_e}$& Observed visual information (bag of features)\\
$w_{t_e}$& Observed linguistic information (bag of words)\\
$R_{t_e}$& Index of a spatial region\\
$C_{t_e}$& Index of a spatial concept\\
$\pi_{e,l}$& Parameter of categorical distribution for region $R_{t_e}$\\
$\mu_{e,m},\Sigma_{e,m}$& Parameters of a Gaussian distribution\\
$G_0$& Parameter of Dirichlet process for $G_e$\\ 
$G_e$& Parameter of the categorical distribution for $C_{t_e}$\\ 
$\phi^v_l$& Parameter of the multinomial distribution for visual information \\
$\phi^w_l$& Parameter of the multinomial distribution for linguistic information\\ 
$\theta^v_{e,l}$& Parameter of multinomial distribution for visual information in environment $e$\\ 
$\theta^w_{e,l}$& Parameter of the multinomial distribution for linguistic information in an environment $e$\\ 
$\mu_{0},\kappa_0$ $\Psi_0,\nu_0$& Hyperparameter of a Gaussian Wishart distribution\\
$\beta$& Hyperparameter of a Dirichlet distribution for $\pi_{e,l}$\\ 
$\alpha^v$& Hyperparameter of Dirichlet distribution for $\phi^v_l$\\ 
$\alpha^w$& Hyperparameter of Dirichlet distribution for $\phi^w_l$\\ 
$\delta^v$& Hyperparameter of a multinomial distribution for $\theta^v_{e,l}$\\ 
$\delta^w$& Hyperparameter of a multinomial distribution for $\theta^w_{e,l}$\\ 
$\gamma,\gamma_0$& Hyperparameters of a Dirichlet process for $G_e, G_0$\\ 
\botrule
\end{tabular}
}
\label{model_val3}
\end{center}
\end{table}

\subsection{Generative process of the proposed model}
The generative process of the proposed model is described as follows:

As the generative process of a category $C_{t_e}$,
\begin{flalign}
G_0  &\sim \mathrm{GEM}(\gamma_0) \label{gene1}\\
G_e &\sim \mathrm{DP}(G_{0},\gamma) \label{gene2}\\
C_{t_e} &\sim \mathrm{Cat}(G_e), \label{gene3}
\end{flalign}
As the generative process of observations $v_{t_e}$ and $w_{t_e}$,
\begin{flalign}
\phi^v_l &\sim \mathrm{Dir}(\alpha^{v}) \label{gene4}\\
\phi^w_l &\sim \mathrm{Dir}(\alpha^{w}) \label{gene5}\\
\theta^v_{e,l} &\sim \mathrm{Dir}(\phi^v_l, \delta^v) \label{gene6}\\
\theta^w_{e,l} &\sim \mathrm{Dir}(\phi^w_l, \delta^w) \label{gene7}\\
v_{t_e} &\sim \mathrm{Mult}(\theta^v_{e,C_{t_e}}) \label{gene8}\\
w_{t_e} &\sim \mathrm{Mult}(\theta^w_{e,C_{t_e}}), \label{gene9}
\end{flalign}
As the generative process of a region $R_{t_e}$,
\begin{flalign}
x_{t_e} &\sim p(x_{t_e} \mid x_{t-1_e},u_{t_e}) \label{gene10}\\
z_{t_e} &\sim p(z_{t_e} \mid x_{t_e}) \label{gene11}\\
\pi_{e,l} &\sim \mathrm{GEM}(\beta) \label{gene12}\\
\Sigma_{e,m} &\sim \mathcal{IW}(\Psi_0,\nu_0) \label{gene13}\\
\mu_{e,m} &\sim \mathcal{N}(\mu_0,\Sigma_{e,m}/\kappa_0) \label{gene14}\\
R_{t_e} &\sim p(R_{t_e} \mid x_{t_e},C_{t_e},\bm{\mu_e},\bm{\Sigma_e},\bm{\pi_e}), \label{gene15}
\end{flalign}
where $\mathrm{GEM}(\cdot)$ is a GEM distribution, $\mathrm{DP}$ is a Dirichlet process, $\mathrm{Cat}$ is a categorical distribution, $\mathrm{Dir}(\cdot)$ is a Dirichlet distribution, $\mathrm{Mult}(\cdot)$ is a multinomial distribution, $\mathcal{IW}(\cdot)$ is the inverse Wishart distribution, and $\mathcal{N}(\cdot)$ is a Gaussian distribution.

Detailed definitions of the generative process of the proposed model are described in Appendix~\ref{apdx:proposed_method:generate}.

\subsection{Acquisition of multimodal information}

\subsubsection{Positional information}
MCL~\cite{MCL} is used to acquire the position information. Position estimation is performed based on a map created in advance by SLAM~\cite{SLAM}. Specifically, the position distribution $p(x_t \mid x_{t-1}, z_t, u_t)$ is estimated based on the observed information $z_t$ at time $t$ and the control information $u_t$. The positional information $x_t$ is obtained using maximum a posteriori probability estimation as follows:
\begin{equation}
\label{eq:position}
x_t = \argmax_{x_t}p(x_t \mid x_{0},z_{1:t},u_{1:t}).
\end{equation}

\subsubsection{Visual information}
An image captured by a camera attached to a robot is converted to a visual feature as an observation using Caffe~\citep{caffe}, which is a framework for CNNs~\citep{CNN} provided by the Berkeley Vision and Learning Center.
The parameters of CNNs were trained using the dataset from the ImageNet Large Scale Visual Recognition Challenge 2012\footnote{ILSVRC2012: http://www.image-net.org/challenges/LSVRC/2012/}.
Visual information $v_{t} = [v_{t,1},v_{t,2},\cdots,v_{t,I}]$ is calculated using the following equation:
\begin{equation}
\label{eq:vision}
v_{t} = f_{t} \times S^v,
\end{equation}
where $f_t = [f_{t,1},f_{t,2},\cdots,f_{t,I}]$ are the output values of the units in the $7_{th}$ layer of Caffe. $I$ is the number of output units at the $7_{th}$ layer and was set to $4096$ in the experiment. We selected the $7_{th}$ layer for use in Caffe as a visual feature extractor, not as an object's label recognizer. $S^v$ denotes an increment parameter for converting the output of the units in Caffe into a 
bag-of-feature representation. In the experiment, $S^v$ was empirically set to $5.0$. 

\subsubsection{Linguistic information}
A linguistic instruction obtained from the user at time $t$ is converted to a word sequence $L_t = [L_{t,1},\cdot \cdot \cdot, L_{t,J}]$. $J$ denotes the number of words in the linguistic instruction. 
A dictionary $D = [D_1, \cdot \cdot \cdot, D_K]$ is generated from a set of obtained word sequences $L_t$ in a dataset. $K$ denotes the number of word types in the dictionary.
The linguistic information $w_{t,k} \in \{w_{t,1},w_{t,2},\cdots,w_{t,K}\}$ at time $t$ as a bag-of-words representation is calculated as follows:
\begin{flalign}
\label{eq:word}
w_{t,k}= 
\begin{cases}
    S^w & (L_{t,j}=D_k) \\
    0 & (\text{otherwise}),
  \end{cases}
\end{flalign}
where $S^w$ denotes an increment parameter for linguistic information.
In the experiment, $S^w$ was empirically set to $5.0 \times 10^3$. $S^v$ and $S^w$ are empirically determined by considering the balance of the number of dimensions between the modalities.

\subsection{Inference process}
\label{sec:Inference process}
During the learning of spatial concepts, the robot estimates the set of all the latent variables, $\mathbf{V}=\{ C_{i}, R_{i} \}_{i=1}^{I}$, and the set of model parameters, $\Theta = \{ \mu_{e,m}, \Sigma_{e,m}, \pi_{e,l}, \phi^{v}_{l}, \theta^v_{e,l}, \theta^w_{e,l}, \phi^{v}_{l}, \phi^{w}_{l}, G_{e}, G_{0} \}$, from the set of multimodal observations, $\mathbf{O} = \{ x_{i}, v_{i}, w_{i} \}_{i=1}^{I}$ through Gibbs sampling.
$I$ is the number of data points. The sampling values are provided by the iteration of Gibbs sampling from the joint posterior distribution as follows:
\begin{eqnarray}
  \mathbf{V}, \Theta  \sim p \left(\mathbf{V}, \Theta \mid\mathbf{O}, \mathbf{h} \right)
  \label{eq:gibbs} 
\end{eqnarray}
where the set of hyperparameters is denoted as $\mathbf{h} = \{ \alpha^v, \alpha^w, \beta, \gamma_0, \gamma, \mu_{0}, \kappa_{0}, \psi_{0}, \nu_{0}, \delta^v, \delta^w \}$.
We provide the details of the Gibbs sampling in Appendix~\ref{apdx:proposed_method:gibbs} and the algorithm of the inference process in Appendix~\ref{apdx:proposed_method:algorithm}.

\subsection{Prediction of location names} \label{n_t_est}
The proposed method enables a robot to predict a location name from the observed positional information, visual information, and estimated model parameters using the following formula:
\begin{flalign}
\hat{w}_{t_e} =&\argmax_{w_{t_e}} \{ p(w_{t_e} \mid x_{t_e},v_{t_e})\times I(w_{t_e};C_{t_e} \mid \Theta) \}, 
\end{flalign}
where $\hat{w_{t_e}}$ denotes the predicted location name, $p(w_{t_e} \mid x_{t_e},v_{t_e})$ denotes the probability of the predicted location names, and $I(w_{t_e};C_{t_e} \mid \Theta)$ denotes the mutual information between $C_{t_e}$ and $w_{t_e}$.

The probability of the predicted location names $p(w_{t_e} \mid x_{t_e},v_{t_e})$ is calculated using the following formulas:
\begin{flalign}
p(w_{t_e} \mid x_{t_e},v_{t_e}) =&\sum_{C_{t_e}}p(w_{t_e} \mid C_{t_e})\sum_{R_{t_e}}p(C_{t_e},R_{t_e} \mid x_{t_e},v_{t_e}) \nonumber \\
\propto&\sum_{C_{t_e}}\sum_{R_{t_e}} \Bigl[ p(w_{t_e} \mid \theta^w_{e,C_{t_e}})p(v_{t_e} \mid \theta^v_{e,C_{t_e}})p(C_{t_e} \mid G_e) \Bigr. \nonumber \\
&\times \Bigl. p(x_{t_e} \mid \mu_{R_{t_e}},\Sigma_{R_{t_e}})p(R_{t_e} \mid \pi_{e,C_{t_e}}) \Bigr].
\label{eq:name_prediction}
\end{flalign}

Linguistic instructions from the user about a place are given to the robot as sentences that also include words that do not represent a place, such as `the,' `here,' and `is.'
To reduce the effect of these words on the prediction of location names, we used mutual information (MI) between a spatial concept and a word.
The mutual information between a spatial concept and a word $I(w_{t_e};C_{t_e}\mid\Theta)$ is calculated using the following formula:
\begin{flalign}
I(w_{t_e};C_{t_e} \mid \Theta)=\sum_w \sum_cP(w,c \mid \Theta)\log\frac{P(w,c \mid \Theta)}{P(w \mid \Theta)P(c \mid \Theta)},
\label{eq:MI}
\end{flalign}
where $\Theta$ denotes the set of estimated model parameters $\{$ $\mu_{e,m}$, $\Sigma_{e,m}$, $\pi_{e,l}$, $R_{t_e}$, $\theta^v_{e,l}$, $\theta^w_{e,l}$, $G_e$, $\phi^v_l$, $\phi^w_l$, $G_0$ $\}$, $w\in(w_t,\overline{w_t})$ and $\overline{w_t}$ denote a set of linguistic information other than $w_t$, and $c \in (C_t,\overline{C_t})$ and $\overline{C_t}$ denote a set of spatial concepts other than $C_t$.
Since the words that represent places are used only for instructions in a place, the amount of mutual information becomes large. In contrast, since the words that do not represent places are used for instructions in various places, the amount of mutual information becomes small.

\subsection{Prediction of positions} \label{x_t_est}
In the proposed method, a position indicated by a location name can be predicted from the observed linguistic information and estimated model parameters using the following formula:
\begin{flalign}
\hat{x}_{t_e} &\sim \mathcal{N}(x_{t_e} \mid  \mu_{e,\hat{R}_{t_e}},\Sigma_{e,\hat{R}_{t_e}}),
\label{eq:position_prediction}
\end{flalign}
where the predicted position ($\hat{x}_{t_e}$) in which the robot moves can be sampled based on the model parameters of the Gaussian distribution with a predicted region ($\hat{R_{t_e}}$).

$\hat{R_{t_e}}$ is calculated using the following formula:
\begin{flalign}
\hat{R_{t_e}} &=\argmax_{R_{t_e}}p(R_{t_e} \mid w_{t_e}), \nonumber \\
\end{flalign}
where $p(R_{t_e} \mid w_{t_e})$ denotes the probability of the predicted regions calculated using the following formulas:
\begin{flalign}
p(R_{t_e} \mid w_{t_e}) &= \sum_{C_{t_e}} p(R_{t_e} \mid C_{t_e})p(C_{t_e} \mid w_{t_e}) \nonumber \\
&\propto \sum_{C_{t_e}} p(R_{t_e} \mid \pi_{e,C_{t_e}})p(w_{t_e} \mid \theta^w_{e,C_{t_e}})p(C_{t_e} \mid G_e).
\end{flalign}

To evaluate the learning of a spatial region on the meaning of a location name, we used the positions sampled from the Gaussian distribution rather than the mean of the Gaussian distribution.

\section{Experiment for the generalization of spatial concepts to environments}
\subsection{Overview}
When a human-support robot is delivered to a new home environment, the robot must linguistically explain its self-position, such as `I am in the kitchen,' without the user’s linguistic instruction. Furthermore, the robot must move to a suitable place according to the command `come to the living room' without any instruction from the user.
Assuming such tasks, we performed an experiment to evaluate the performance of the proposed model for the generalization of spatial concepts to environments based on the tasks of name and position prediction. The experiment was conducted in nineteen virtual home environments consisting of sixteen experienced environments, one validation environment, and three new environments. In this experiment, the targets of evaluation were general places that can be transferred between environments such as `kitchen' and `living room.' The performance of the proposed model was evaluated in terms of the accuracy of the name prediction from position and vision information, and the position prediction from location names by comparing it with conventional spatial concept models. Additionally, a transition in the accuracy of name and position prediction in the proposed model was observed when the number of experienced environments changed.

\subsection{Experimental condition}
\subsubsection{Dataset}\label{sec:Dataset}
\begin{figure}[tb]
  \begin{center}
    \includegraphics[width=400pt]{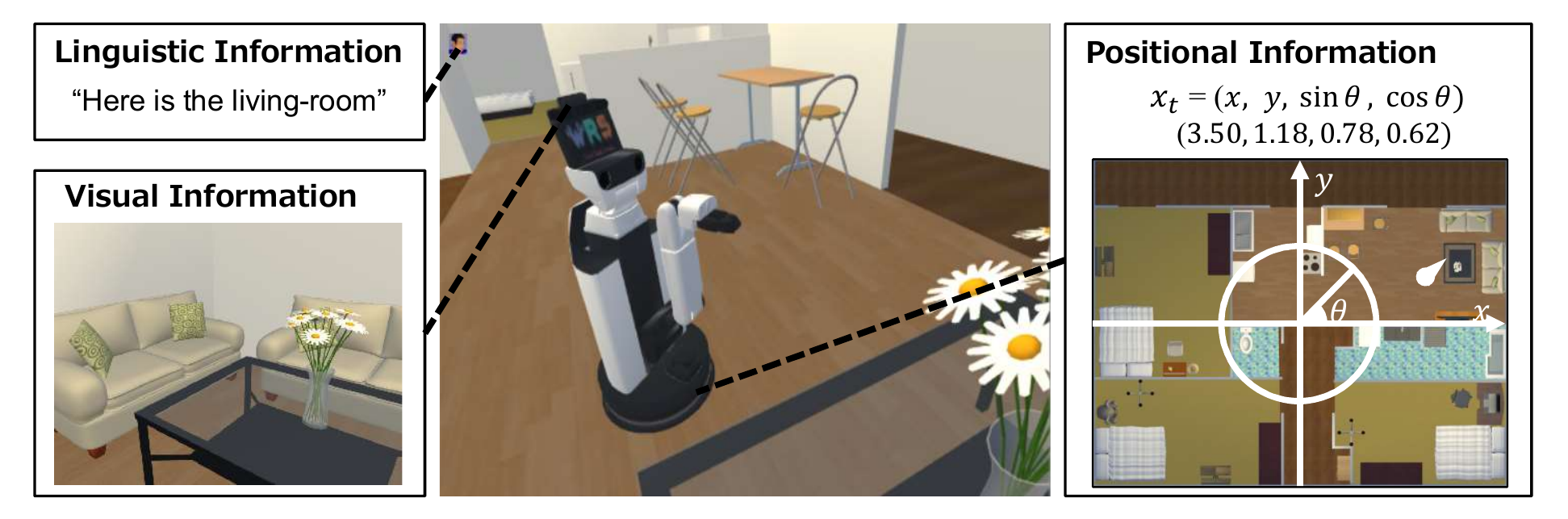}
    \caption{Simulator to collect multimodal information: a scene in which a robot obtains visual, positional, and linguistic information in a virtual home environment using SIGVerse.}
   \label{sigverse}
  \end{center}
\end{figure}
\begin{figure}[tb]
  \begin{center}
    \includegraphics[width=400pt]{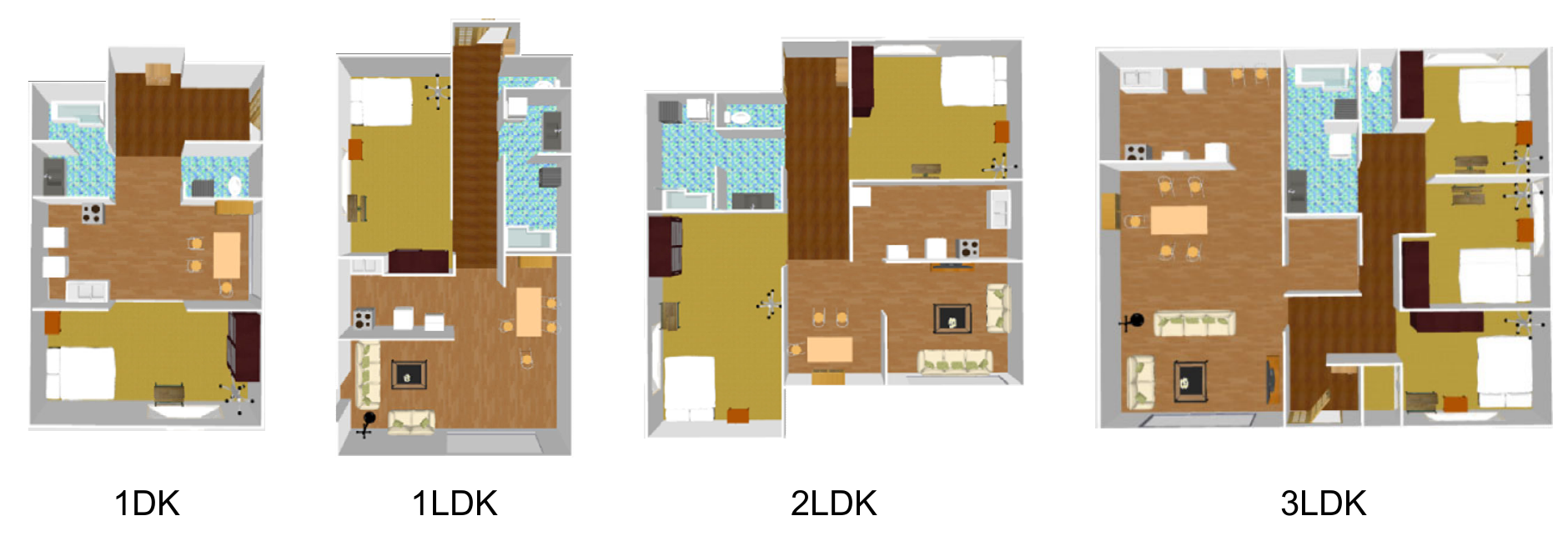}
    \caption{Experimental environments: examples of four types of virtual home environments, i.e., 1DK (1 room, dining, and kitchen), 1LDK (1 room, living, dining, and kitchen), 2LDK (2 rooms, living, dining, and kitchen), and 3LDK (3 rooms, living, dining, and kitchen).}
   \label{environments}
  \end{center}
\end{figure}
We prepared twenty-three home environments for the experiment using SIGVerse~\cite{SIGVerse}, which is a simulator for obtaining multimodal data through human--robot interaction. Fig.~\ref{sigverse} shows a scene in which a robot acquires visual, positional, and linguistic information in a virtual environment on SIGVerse. Examples of virtual home environments are shown in Fig.~\ref{environments}. The twenty-three home environments consisted of four types: 1DK, 1LDK, 2LDK, and 3LDK. Each type had several different furniture layouts. In the simulator, a human support robot (HSR: a mobile manipulator created by the Toyota company)~\cite{HSR} was used as a robot model to explore an environment based on robot operating system (ROS) packages.
We prepared twenty sets of multimodal information data for each place, e.g., kitchen, entrance, and living room. The multimodal dataset used for the experiment in the twenty-three home environments can be downloaded from a link~\footnote{transferlearning\_spco:  https://github.com/is0314px/transferlearning\_spco/tree/master/source/gibbs\_dataset/sigverse}. The collection process of multimodal information data in SIGVerse is described in Appendix~\ref{apdx:experiment_g:data}.

\begin{table}[tb]
\caption{Example of sentences given as linguistic instructions}
\begin{center}
\scalebox{0.8}{
\begin{tabular}{c p{6cm} c p{6cm}}
\toprule
No. & Example of given sentences & No. & Example of given sentences\\
\colrule
1& This is the kitchen.& 6& Living-room is here.\\
2& Toilet is here.& 7& This is the bedroom.\\
3& This place is the dining.& 8& Bath is here.\\
4& This location is the entrance.& 9& This place is washroom.\\
5& This space is the dining.& 10& This location is the kitchen.\\
\botrule
\end{tabular}
}
\label{tab:sentences}
\end{center}
\end{table}

\subsubsection{Comparison models}
To evaluate the performance for the generalization of spatial concepts to environments, we prepared eight models as baseline and comparison models, i.e., SpCoA, SpCoA+MI, and the proposed model with 0, 2, 3, 4, 8, and 16 experienced environments. The details of the comparison models are described in Appendix~\ref{apdx:experiment_g:models}. 

\subsubsection{Inference of model parameters from the dataset}
We divided the dataset in Sec.~\ref{sec:Dataset} to nineteen experienced environments, one validation environment, and three new environments. To evaluate the performance of the comparison models, three new environments were used to evaluate the performance of name and position prediction using the comparison models based on the ground truth provided by the user. 
The details of the inference of the model parameters from the dataset in comparison models are described in Appendix~\ref{apdx:experiment_g:inference}. 

\subsubsection{Evaluation criteria}
We evaluated the performance of the proposed model based on the prediction accuracy in tasks of name and position prediction based on the ground truth provided by the user. 
Fig.~\ref{grobal_position_pre} (a) shows the example of the ground truth for regions corresponding to location names in a new environment. The area of each place was defined as a rectangle region by the user.
The details of evaluation criteria in name and position prediction are described in Appendix~\ref{apdx:experiment_g:criteria}.

\subsection{Experimental result}

\begin{figure}[tb]
  \begin{center}
    \begin{tabular}{c}
      \begin{minipage}{0.45\hsize}
        \begin{center}
          \includegraphics[clip, width=7.0cm]{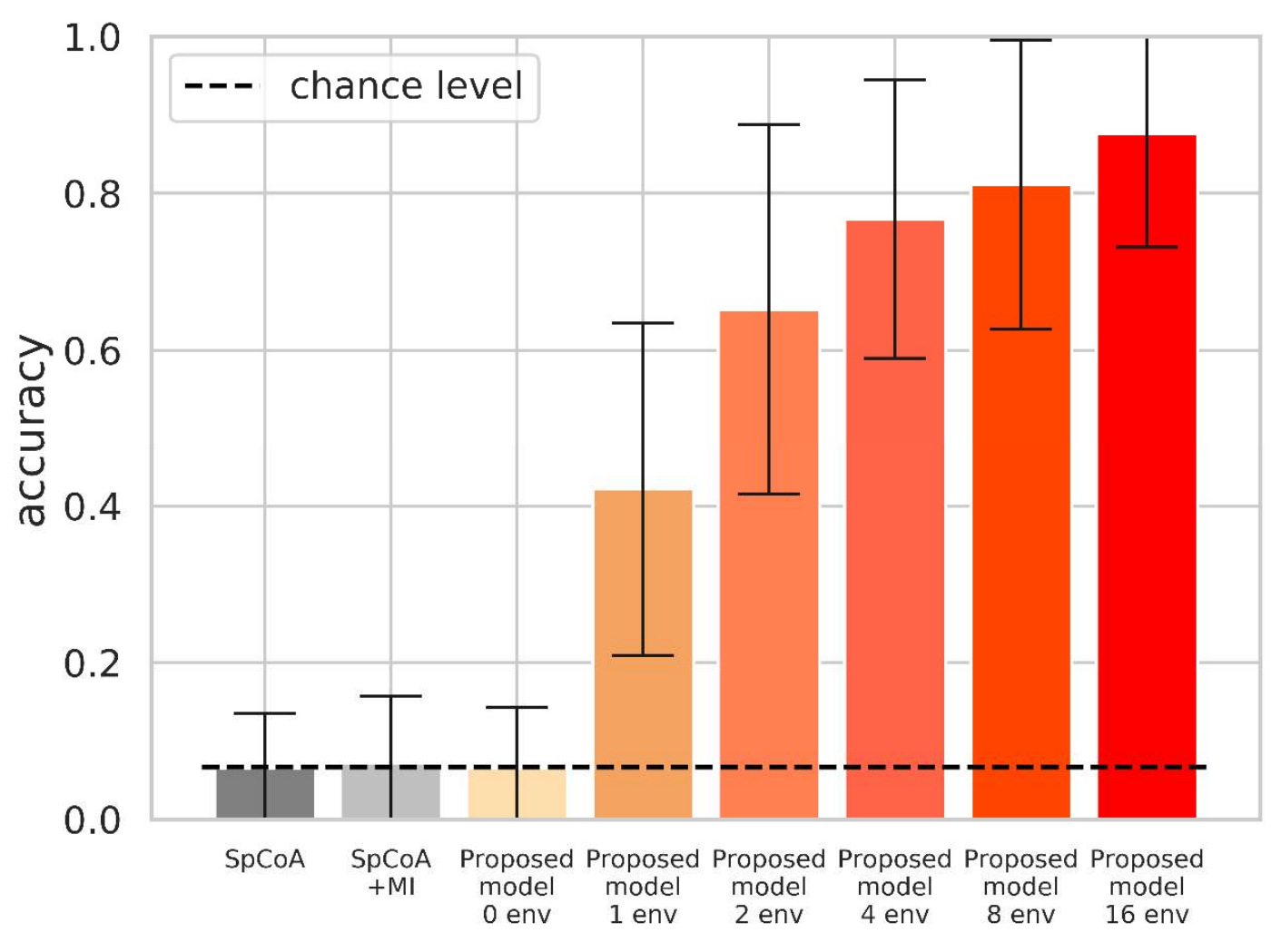}
          \hspace{1.6cm} (a)
        \end{center}
      \end{minipage}
      \begin{minipage}{0.45\hsize}
        \begin{center}
          \includegraphics[clip, width=7.0cm]{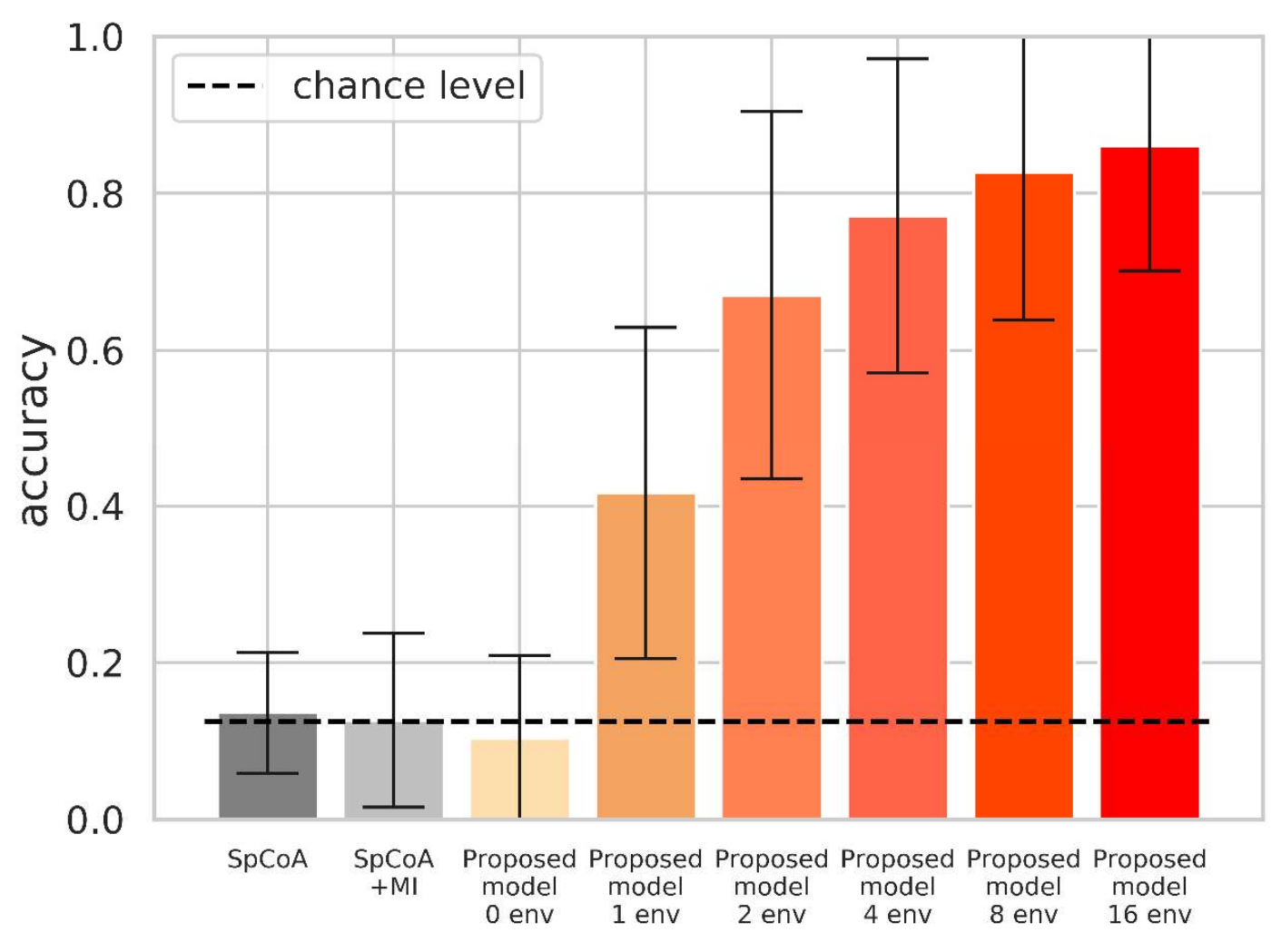}
          \hspace{1.6cm} (b)
        \end{center}
      \end{minipage}
    \end{tabular}
    \caption{Experimental results of name and position prediction by the baseline and proposed models. (a) Accuracy of the predicted location name. (b) Accuracy of the predicted positions. The average and standard deviation are shown in the graphs in (a) and (b).}
    \label{grobal_result}
  \end{center}
\end{figure}

Fig.~\ref{grobal_result} shows the experimental results of name and position prediction by the baseline and proposed models. (a) and (b) show the accuracy of the predicted location names, the accuracy of predicted positions, respectively. The average and standard deviation on twenty inferences through Gibbs sampling for three new environments are shown in the graphs of (a) and (b).
In graph (a) of the name prediction in new environments, the accuracy of SpCoA and SpCoA+ML is the same as the chance level because the name prediction by these models became a random choice from a uniform distribution inferred only from the vision and position information in a new environment. In the results of the proposed models, the accuracy of the predicted location names increased significantly as the number of experienced environments increased from zero to sixteen environments, even in new environments.
In graph (b) of the accuracy of predicted positions in the new environment, SpCoA and SpCoA+ML were at the chance level, but in the proposed model, the accuracy of the predicted positions improved significantly from zero to sixteen experienced environments.

\begin{table}[tb]
\caption{Accuracy of name prediction by SpCoA, SpCoA+MI, and the proposed model for each place. The accuracy is shown as `average (standard deviation)' in twenty trials. The largest values in each place are indicated in bold and underlined. The second largest value in each place is indicated in bold.}
\begin{center}
\scalebox{0.68}{
\begin{tabular}{c c c c c c c c c}
\toprule
Model & Entrance & Living-room & Washroom & Toilet & Dining & Bedroom & Bath & Kitchen\\
\colrule
SpCoA & 0.09 (0.28) & 0.06 (0.25) & 0.05 (0.22) & 0.06 (0.25) & 0.02 (0.13) & 0.05 (0.20) & 0.10 (0.30) & 0.13 (0.34)\\
SpCoA+MI & 0.12 (0.32) & 0.11 (0.31) & 0.03 (0.18) & 0.02 (0.13) & 0.10 (0.29) & 0.08 (0.22) & 0.05 (0.22) & 0.07 (0.25)\\
\colrule
Proposed model\\
0 env. & 0.05 (0.20) & 0.08 (0.26) & 0.11 (0.29) & 0.03 (0.16) & 0.04 (0.18) & 0.05 (0.14) & 0.05 (0.22) & 0.11 (0.31)\\
1 env. & 0.57 (0.49) & 0.42 (0.49) & 0.44 (0.50) & 0.44 (0.50) & 0.33 (0.46) & 0.37 (0.34) & 0.45 (0.50) & 0.44 (0.49)\\
2 env. & 0.82 (0.39) & 0.57 (0.48) & 0.70 (0.46) & 0.80 (0.40) & 0.49 (0.47) & 0.60 (0.33) & 0.68 (0.47) & 0.65 (0.46)\\
4 env. & 0.87 (0.33) & 0.68 (0.46) & \underline{\textbf{0.91 (0.28)}} & 0.81 (0.39) & 0.67 (0.45) & 0.69 (0.34) & 0.80 (0.40) & \underline{\textbf{0.83 (0.35)}}\\
8 env. & \textbf{0.93 (0.25)} & \textbf{0.80 (0.40)} & 0.87 (0.33) & \underline{\textbf{0.89 (0.31)}} & \textbf{0.68 (0.44)} & \textbf{0.74 (0.31)} & \underline{\textbf{0.90 (0.30)}} & \textbf{0.81 (0.37)}\\
16 env. & \underline{\textbf{0.97 (0.16)}} & \underline{\textbf{0.92 (0.26)}} & \textbf{0.89 (0.30)} & \textbf{0.88 (0.32)} & \underline{\textbf{0.84 (0.34)}} & \underline{\textbf{0.84 (0.25)}} & \textbf{0.88 (0.32)} & \textbf{0.81 (0.37)}\\
\botrule
\end{tabular}
}
\label{tab:name_pred}
\end{center}
\end{table}

\begin{table}[tb]
\caption{Accuracy of position prediction by SpCoA, SpCoA+MI, and the proposed model for each location name. The accuracy is shown as `average (standard deviation)' in twenty trials. The largest values in each place are indicated in bold and underlined. The second largest value in each place is indicated in bold.}
\begin{center}
\scalebox{0.68}{
\begin{tabular}{c c c c c c c c c}
\toprule
Model & Entrance & Living-room & Washroom & Toilet & Dining & Bedroom & Bath & Kitchen\\
\colrule
SpCoA & 0.09 (0.28) & 0.06 (0.25) & 0.05 (0.22) & 0.06 (0.25) & 0.02 (0.13) & 0.05 (0.20) & 0.10 (0.30) & 0.13 (0.34)\\
SpCoA+MI & 0.12 (0.32) & 0.11 (0.31) & 0.03 (0.18) & 0.02 (0.13) & 0.10 (0.29) & 0.08 (0.22) & 0.05 (0.22) & 0.07 (0.25)\\
\colrule
Proposed model\\
0 env. & 0.12 (0.32) & 0.18 (0.37) & 0.05 (0.22) & 0.10 (0.30) & 0.06 (0.22) & 0.33 (0.48) & 0.14 (0.34) & 0.12 (0.32)\\
1 env. & 0.55 (0.50) & 0.37 (0.48) & 0.39 (0.49) & 0.37 (0.48) & 0.25 (0.40) & 0.55 (0.50) & 0.42 (0.50) & 0.44 (0.50)\\
2 env. & 0.76 (0.42) & 0.56 (0.48) & 0.71 (0.45) & 0.78 (0.40) & 0.49 (0.46) & 0.80 (0.40) & 0.67 (0.46) & 0.58 (0.49)\\
4 env. & 0.81 (0.39) & 0.63 (0.46) & 0.84 (0.36) & 0.81 (0.39) & \textbf{0.64 (0.43)}& 0.90 (0.30) & 0.73 (0.44) & \underline{\textbf{0.82 (0.39)}}\\
8 env. & \underline{\textbf{0.93 (0.22)}} & \textbf{0.75 (0.40)} & \textbf{0.88 (0.32)} & \textbf{0.83 (0.38)} & 0.62 (0.41) & \underline{\textbf{0.93 (0.25)}} & \underline{\textbf{0.87 (0.32)}} & \textbf{0.80 (0.40)}\\
16 env. & \textbf{0.91 (0.25)} & \underline{\textbf{0.88 (0.27)}} & \underline{\textbf{0.89 (0.30)}} & \underline{\textbf{0.87 (0.32)}} & \underline{\textbf{0.74 (0.35)}} & \textbf{0.92 (0.28)} & \underline{\textbf{0.87 (0.32)}} & \textbf{0.80 (0.40)}\\
\botrule
\end{tabular}
}
\label{tab:pos_pred}
\end{center}
\end{table}
Table~\ref{tab:name_pred} shows the accuracy of name prediction in the test data for each place as the experimental result of name prediction. The accuracy is indicated as average and standard deviation for twenty trials in each place, i.e., entrance, living-room, washroom, toilet, dining, bedroom, bath, and kitchen. We confirmed that the accuracy for all places increased from 0 to 16 experienced environments in the results of the proposed model.
Table~\ref{tab:pos_pred} shows the accuracy of position prediction in the test data for each location name. The accuracy is indicated as the average and standard deviation for 20 trials. We confirmed that the accuracy for all location names increased from 0 to 16 experienced environments in the results of the proposed model.
In some results of name and position prediction, the accuracy decreased slightly even if the number of environments increased, but the difference from the maximum value was small, and we confirmed that the accuracy tended to increase.

\begin{figure}[tb]
  \begin{center}
    \includegraphics[width=420pt]{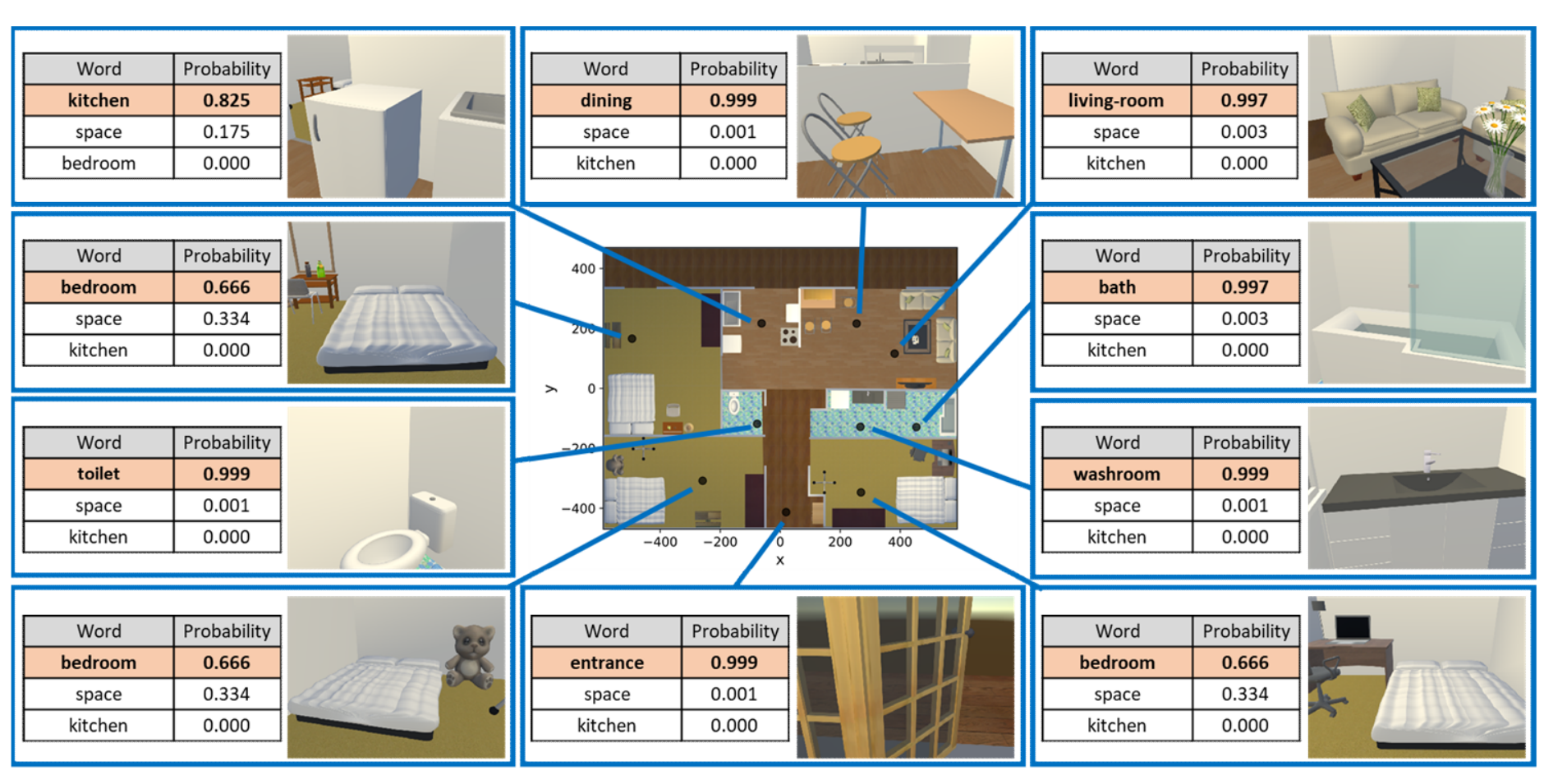}
    \caption{Results of name prediction in a new environment by the proposed model for sixteen experienced environments. The predicted location names are listed as tables of the three best predicted words. The position of the obtained data and the image captured at the position are shown as a black dot on the map and an image in a box, respectively.}
   \label{grobal_name_pre}
  \end{center}
\end{figure}

Fig.~\ref{grobal_name_pre} shows the result of name prediction in a new environment using the proposed model with sixteen experienced environments. The predicted location names are shown as the three best predicted words and probabilities calculated from position and image information using Eq. (\ref{eq:name_prediction}). The black dot in the map and an image in a box shows a position in which the data obtained and an image captured at the position, respectively. For example, the table on the top left confirms that the generalized location name `kitchen' was predicted with the high probability of 0.8251. In the experimental result, general location names were successfully predicted as high probabilities without linguistic instructions from users in a new environment using the proposed model, which transferred the knowledge of spatial concepts formed in experienced environments. 

\begin{figure}[tb]
  \begin{center}
    \begin{tabular}{c}
      \begin{minipage}{0.45\hsize}
        \begin{center}
          \includegraphics[clip, width=7.0cm]{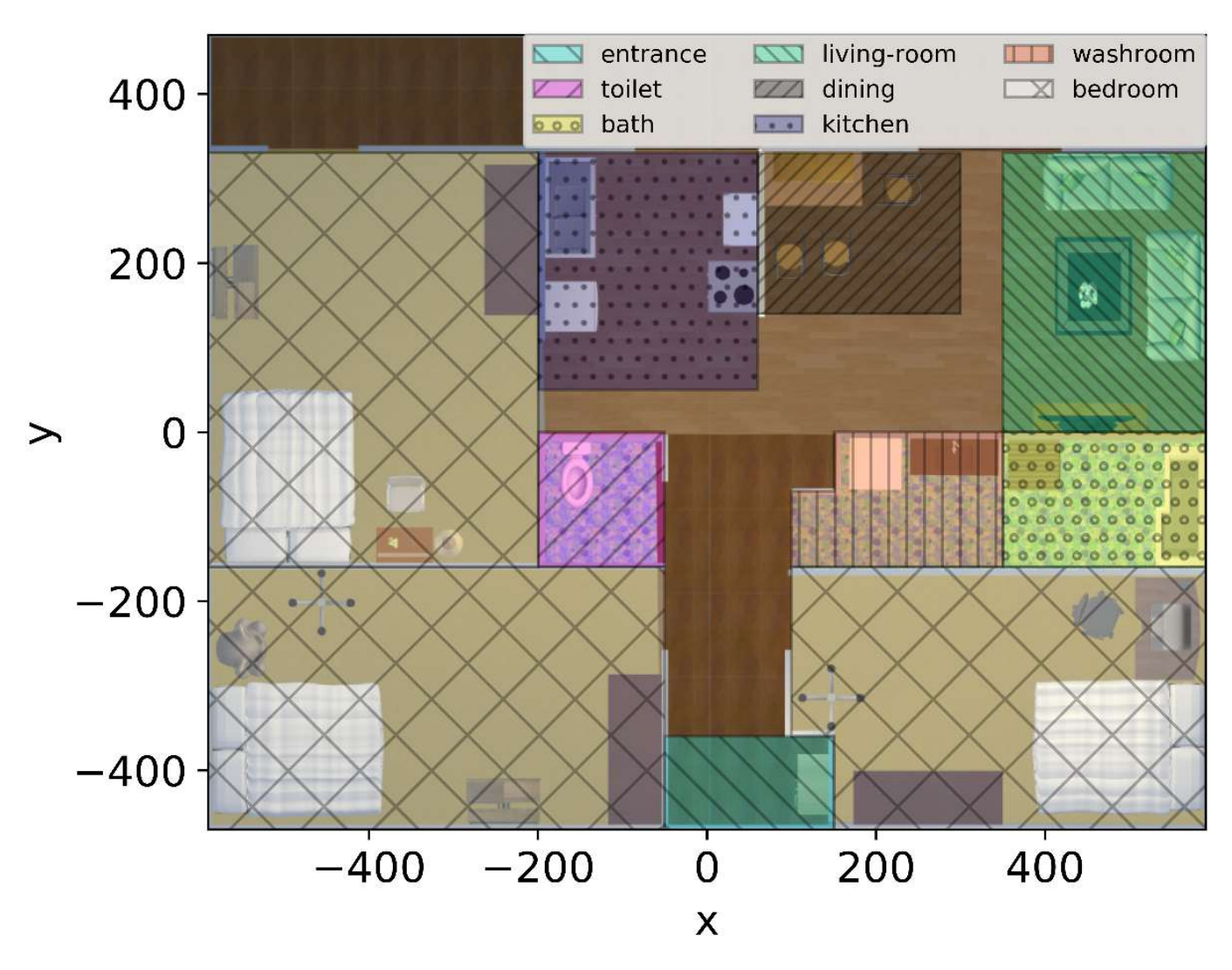}
          \hspace{1.6cm} (a)
        \end{center}
      \end{minipage}
      \begin{minipage}{0.45\hsize}
        \begin{center}
          \includegraphics[clip, width=7.0cm]{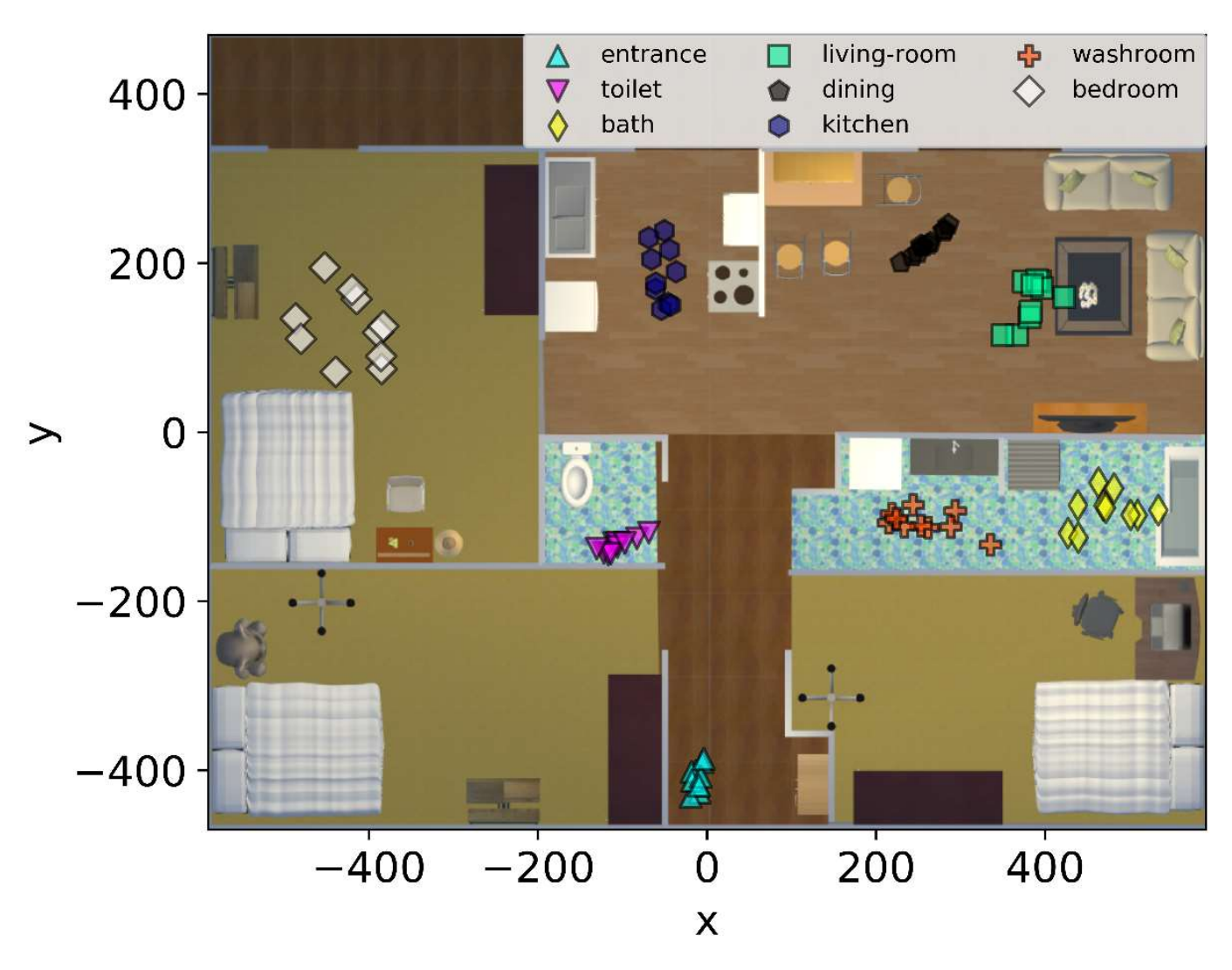}
          \hspace{1.6cm} (b)
        \end{center}
      \end{minipage}
    \end{tabular}
    \caption{(a) Ground truth of regions in a new environment for location names, i.e., entrance, living-room, washroom, toilet, dining, bedroom, bath, and kitchen. (b) Results of position prediction in the new environment using the proposed model with sixteen experienced environments.}
    \label{grobal_position_pre}
  \end{center}
\end{figure}
Fig.~\ref{grobal_position_pre} (a) shows the ground truth of regions in a new environment for location names, i.e., entrance, living-room, washroom, toilet, dining, bedroom, bath, and kitchen, and (b) shows the result of position prediction in the new environment using the proposed model with sixteen experienced environments. In the result of position prediction, a mark shows a predicted position from word information, e.g., entrance, bedroom, and kitchen using Eq. (\ref{eq:position_prediction}). The position was predicted ten times for each location name. We confirmed that the proposed model enables a robot to predict suitable positions from general location names in a new environment without linguistic instructions from users by transferring the knowledge of spatial concepts formed in experienced environments.

\section{Experiment on the adaptive learning of spatial concepts for home-specific places}
\subsection{Overview}
After a human-support robot is delivered to a new home environment, it must lean home-specific places, such as Emma's room and father's room, based on linguistic instructions from the user on site, and utilize them to provide several services. For example, the robot must linguistically explain its position using home-specific location names learned on site such as `Emma's room.' In addition, the robot must move to home-specific places called father's room based on linguistic commands such as `Go to father's room.'
Assuming such tasks, we performed an experiment to evaluate the performance of the proposed model for the adaptive learning of spatial concepts to home-specific places based on the tasks of name and position prediction. The experiment was conducted in eighteen virtual home environments consisting of sixteen experienced environments, one validation environment, and one new environment. In this experiment, the targets of evaluation were home-specific places that could not be transferred between environments such as Emma's room and father's room. The performance of the proposed model was evaluated in terms of the accuracy of the name prediction from position and vision information, and the position prediction from a location name. The accuracy of name and position prediction was indicated with the name given rate for a place in a new environment in the proposed model.
\subsection{Experimental condition}
\subsubsection{Dataset}
\begin{figure}[tb]
  \begin{center}
    \includegraphics[width=400pt]{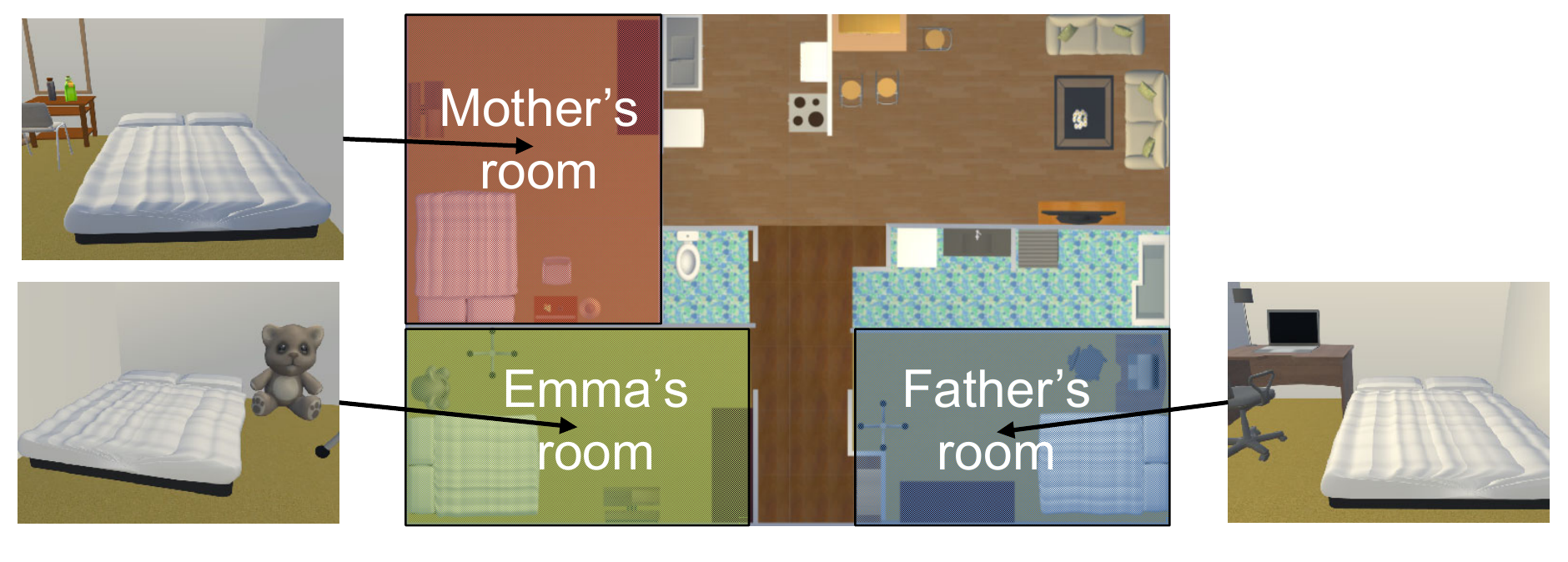}
    \caption{Home specific places, i.e., Emma's room, father's room, and mother's room in a new environment. The three pictures show images captured by a robot in each room.}
   \label{specific_rooms}
  \end{center}
\end{figure}

We prepared a new environment for home-specific places, i.e., Emma's room, father's room, and mother's room, to evaluate the performance of the proposed model in the adaptive learning of spatial concepts for home-specific places. A new environment with home-specific places is shown in Fig.~\ref{specific_rooms}). Three location names, i.e., Emma's room, father's room, and mother's room, were added to the linguistic instructions from the user in the new environment using the same sentences as in Table~\ref{tab:sentences}. For the experienced environments of the proposed model, we used the same dataset as in Section~\ref{sec:Dataset}, which consisted of places representing general location names, e.g., kitchen. The structure of the multimodal data was also the same as the dataset in Section~\ref{sec:Dataset}. 
\subsubsection{Inference of model parameters from the dataset}
Model parameters were trained using a dataset of multimodal information in sixteen experienced environments and a new environment. In the experienced environments, linguistic instructions from the user were given for all the data in every place as linguistic information. In a new environment, linguistic instructions from users were not given for general places such as living rooms, but only to home-specific places such as Emma's room as information for adaptive learning.
To evaluate the adaptive learning process and accuracy of the proposed model for home-specific places in the new environment, the proposed model was trained using a dataset in which the ratio of given location names to forty data points in a home-specific place was set at six levels: $0\%(0/40)$, $20\%(8/40)$, $40\%(16/40)$, $60\%(24/40)$, $80\%(32/40)$, and $100\%(40/40)$.
The hyperparameters of the proposed model were the same as those in Section~\ref{sec:Param}). The inference of the model parameters was performed twenty times for each dataset using Gibbs sampling.
\subsubsection{Evaluation criteria}
The adaptability of the proposed model in the new environment was evaluated based on its ability to predict the location names and positions of home-specific places in the a environment. The inference of model parameters was performed twenty times for 120 training images in three places: Emma's room, father's room, and mother's room in the new environment. The performance of the name and position prediction was evaluated in terms of the accuracy calculated using Eq.~(\ref{eq:name_acc}) and (\ref{eq:pos_acc}) to the ground truth of the sixty test data for the three locations. The accuracy of name and position prediction were indicated using means and confidence intervals for 1200 predicted results consisting of sixty test data with twenty models inferred through Gibbs sampling.

\subsection{Experimental result}
\begin{figure}[tb]
  \begin{center}
    \begin{tabular}{c}
      \begin{minipage}{0.45\hsize}
        \begin{center}
          \includegraphics[clip, width=7.0cm]{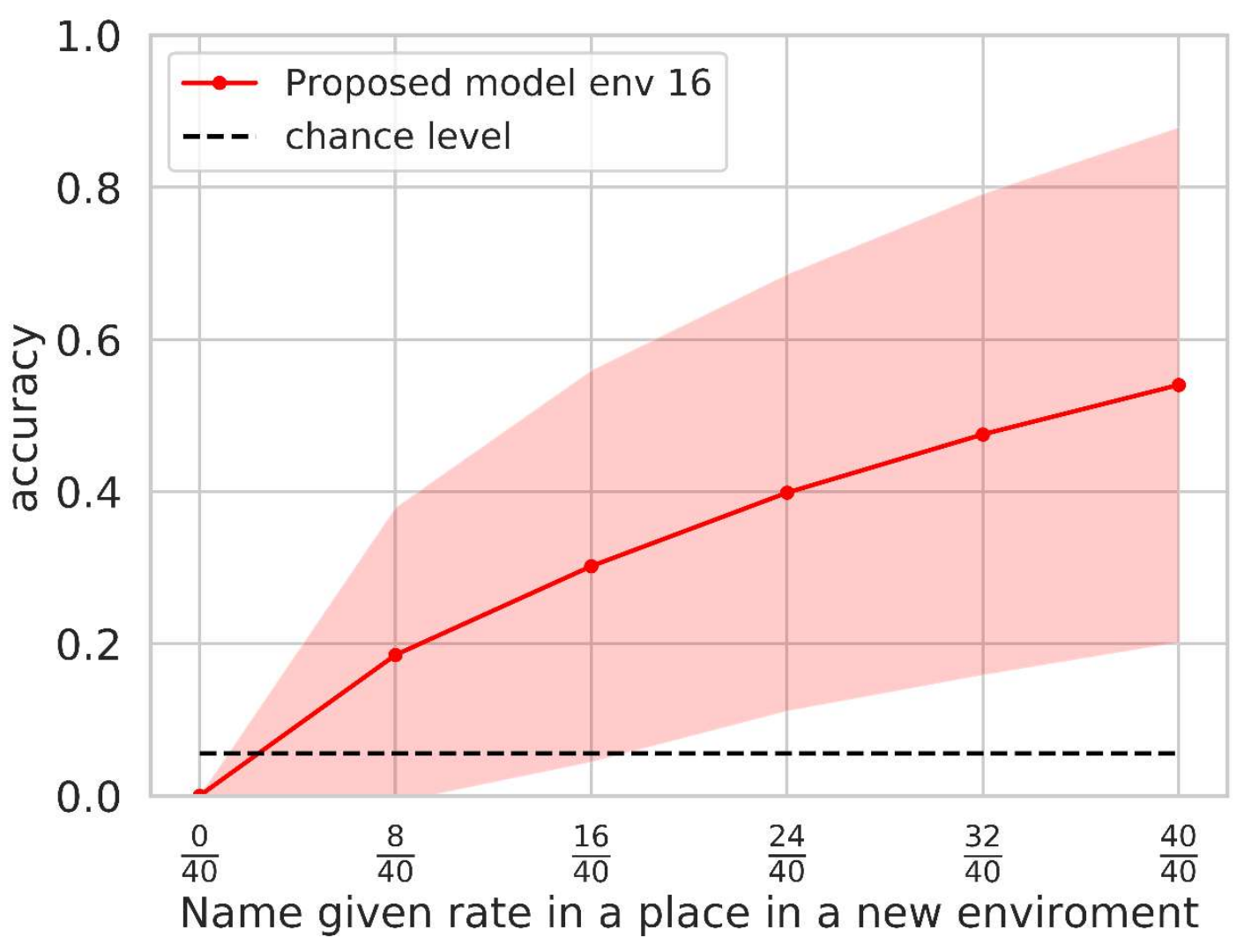}
          \hspace{1.6cm} (a)
        \end{center}
      \end{minipage}
      \begin{minipage}{0.45\hsize}
        \begin{center}
          \includegraphics[clip, width=7.0cm]{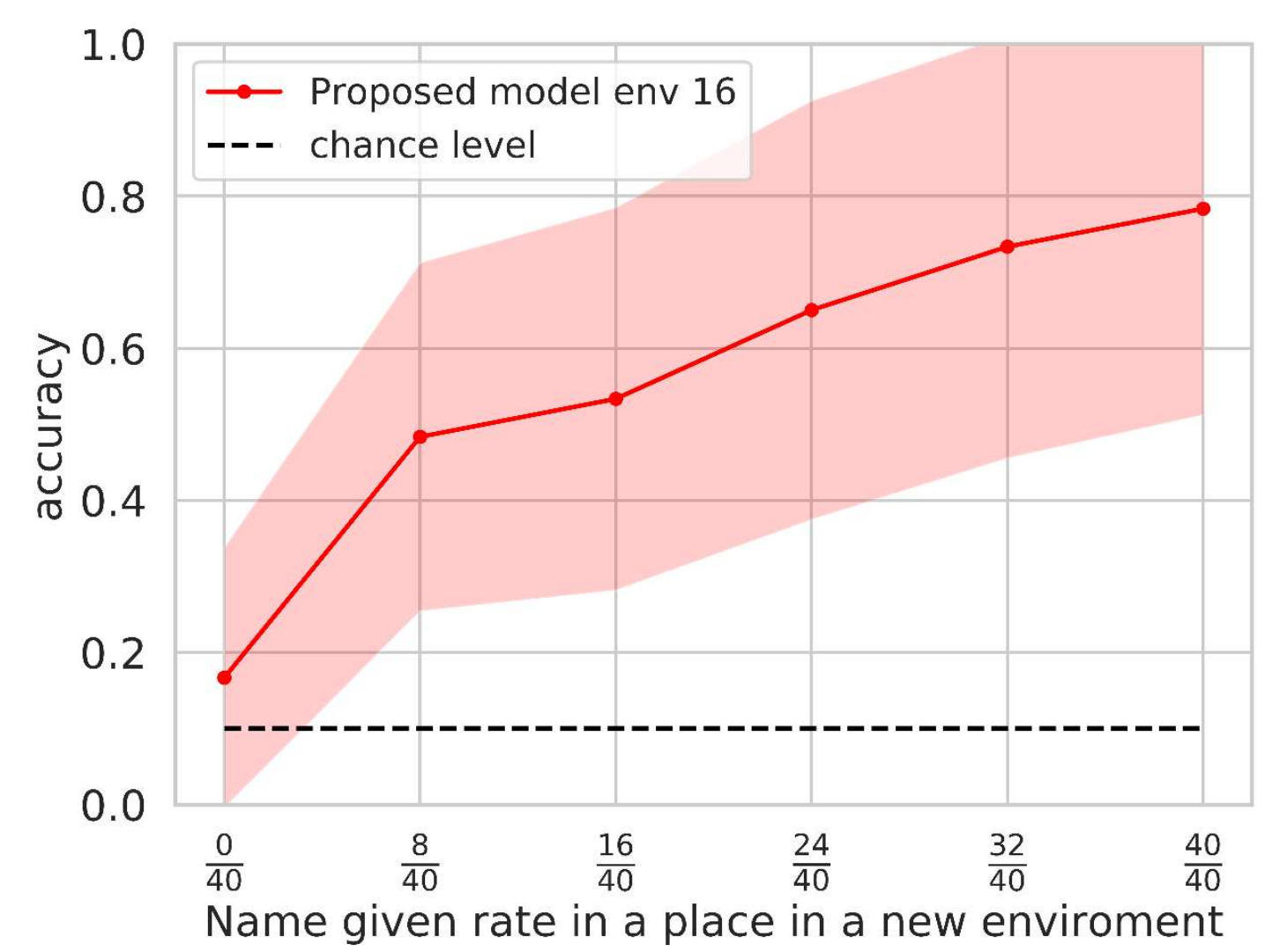}
          \hspace{1.6cm} (b)
        \end{center}
      \end{minipage}
    \end{tabular}
    \caption{Experimental results of name and position prediction for home-specific places using the proposed model trained with sixteen environments. (a) Accuracy of the predicted location name. (b) Accuracy of the predicted positions. The average and standard deviation are shown in the graphs in (a) and (b).}
    \label{local_result}
  \end{center}
\end{figure}
Fig.~\ref{local_result} shows the experimental results of the quantitative evaluation of adaptive learning of spatial concepts in home-specific places using the proposed model trained with sixteen environments. (a) and (b) show the accuracy of name prediction and accuracy of position prediction, respectively.
In the graphs, the horizontal axis shows the name given rate for the data obtained in home-specific places in a new environment, and the vertical axis shows the accuracy of name and position prediction. The name given rate refers to the rate at which the user provided instructions on the location name of 40 data points obtained in home-specific places. The graphs confirm that the accuracy of name and position prediction increased as the name given rate increased for home specific places, i.e., Emma's room, mother's room, and father's room. 

\begin{table}[tb]
\begin{minipage}[t]{.45\textwidth}
\begin{center}
\scalebox{0.70}{
\begin{tabular}{c c c c}
\toprule
Rate & Emma's-room & Mother's-room & Father's-room\\
\colrule
0/40 & 0.00 (0.00) & 0.00 (0.00) & 0.00 (0.00)\\
8/40 & 0.24 (0.36) & 0.19 (0.36) & 0.14 (0.27)\\
16/40 & 0.41 (0.45) & 0.40 (0.46) & 0.10 (0.20)\\
24/40 & 0.41 (0.45) & \textbf{0.44 (0.46)} & 0.35 (0.42)\\
32/40 & \underline{\textbf{0.65 (0.46)}} & 0.43 (0.44) & \textbf{0.35 (0.39)}\\
40/40 & \textbf{0.45 (0.45)} & \underline{\textbf{0.68 (0.46)}} & \underline{\textbf{0.50 (0.43)}}\\
\botrule
\end{tabular}
}
\end{center}
\caption{Accuracy of position prediction by the proposed model for each location name. The accuracy is shown as `average (standard deviation)' in twenty trials. The largest value in each location name is shown in bold and underlined. The second largest value in each location name is shown in bold.}
\label{tab:name_pred_local}
\end{minipage}
\hfill
\begin{minipage}[t]{.45\textwidth}
\begin{center}
\scalebox{0.70}{
\begin{tabular}{c c c c}
\toprule
Rate & Emma's room & Mother's room & Father's room\\
\colrule
0/40 & 0.10 (0.31) & 0.20 (0.41) & 0.20 (0.41)\\
8/40 & 0.65 (0.49) & 0.30 (0.47) & 0.50 (0.51)\\
16/40 & 0.50 (0.51) & 0.65 (0.49) & 0.45 (0.51)\\
24/40 & \underline{\textbf{0.70 (0.47)}} & 0.70 (0.47) & 0.55 (0.51)\\
32/40 & \underline{\textbf{0.70 (0.47)}} & \textbf{0.75 (0.44)} & \textbf{0.75 (0.44)}\\
40/40 & \underline{\textbf{0.70 (0.47)}} & \underline{\textbf{0.85 (0.37)}} & \underline{\textbf{0.80 (0.41)}}\\
\botrule
\end{tabular}
}
\end{center}
\caption{Accuracy of name prediction using the proposed model for each place. The accuracy is shown as `average (standard deviation)' for twenty trials. The largest values in each place are indicated in bold and underlined. The second largest value in each place is indicated in bold.}
\label{tab:pos_pred_local}
\end{minipage}
\end{table}
Table~\ref{tab:name_pred_local} shows the accuracy of name prediction in the test data for each place. The accuracy is indicated as the average and standard deviation for twenty trials. We confirmed that the accuracy for all places increased from zero to nineteen experienced environments in the results of the proposed model.
Table~\ref{tab:pos_pred_local} shows the accuracy of position prediction in the test data for each location name. The accuracy is indicated as the average and standard deviation for twenty trials. We confirmed that the accuracy for all location names increased from zero to nineteen experienced environments in the results of the proposed model.

\begin{figure}[tb]
  \begin{center}
    \includegraphics[width=420pt]{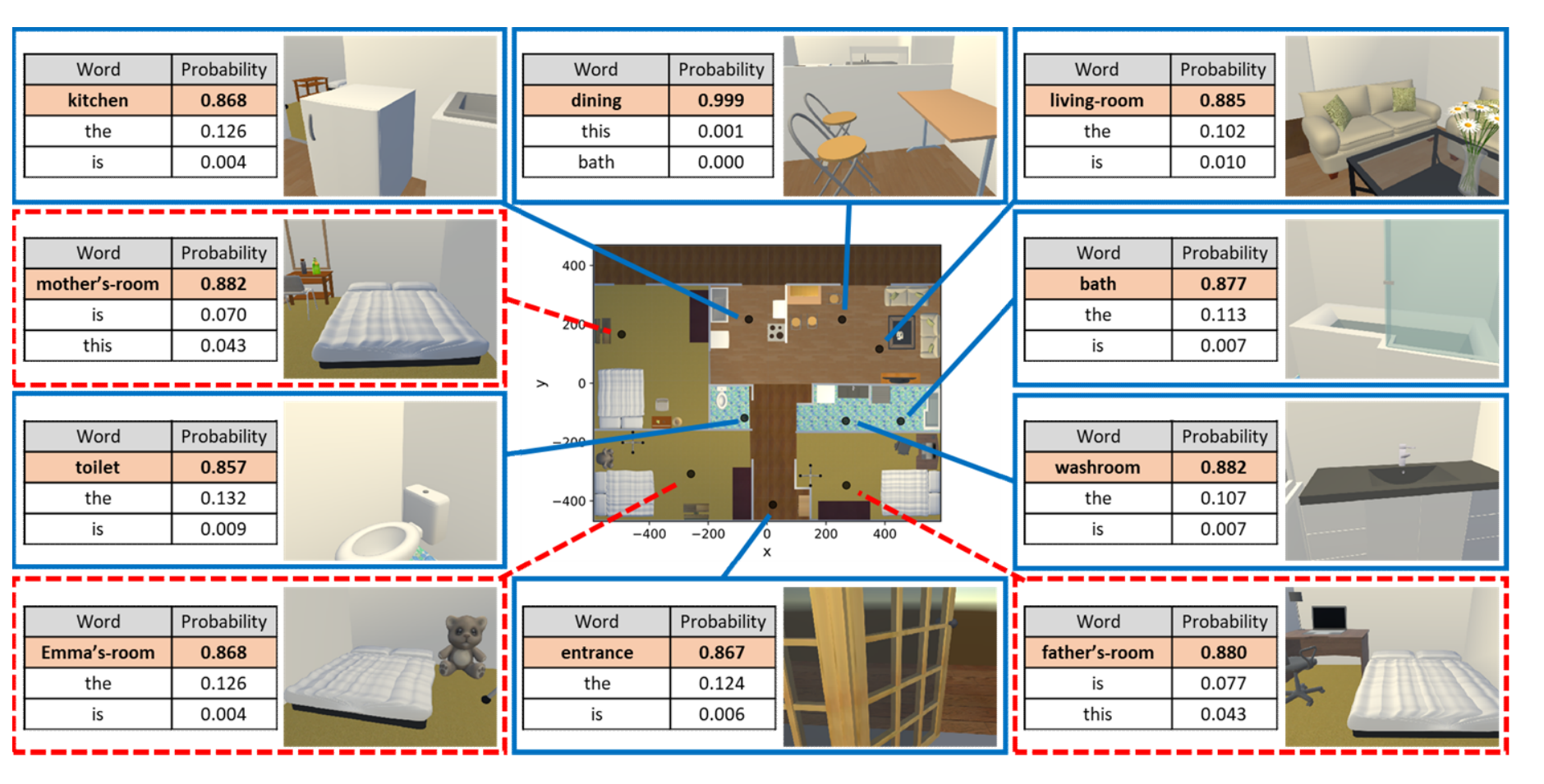}
    \caption{Result of name prediction for home-specific places in a new environment using the proposed model with sixteen experienced environments. The predicted location names are listed as tables of the three best predicted words. The position of the obtained data and image captured at the position are shown as a black bot on the map and an image in a box, respectively.}
   \label{local_name_pre}
  \end{center}
\end{figure}

\begin{figure}[tb]
  \begin{center}
    \begin{tabular}{c}
      \begin{minipage}{0.45\hsize}
        \begin{center}
          \includegraphics[clip, width=7.0cm]{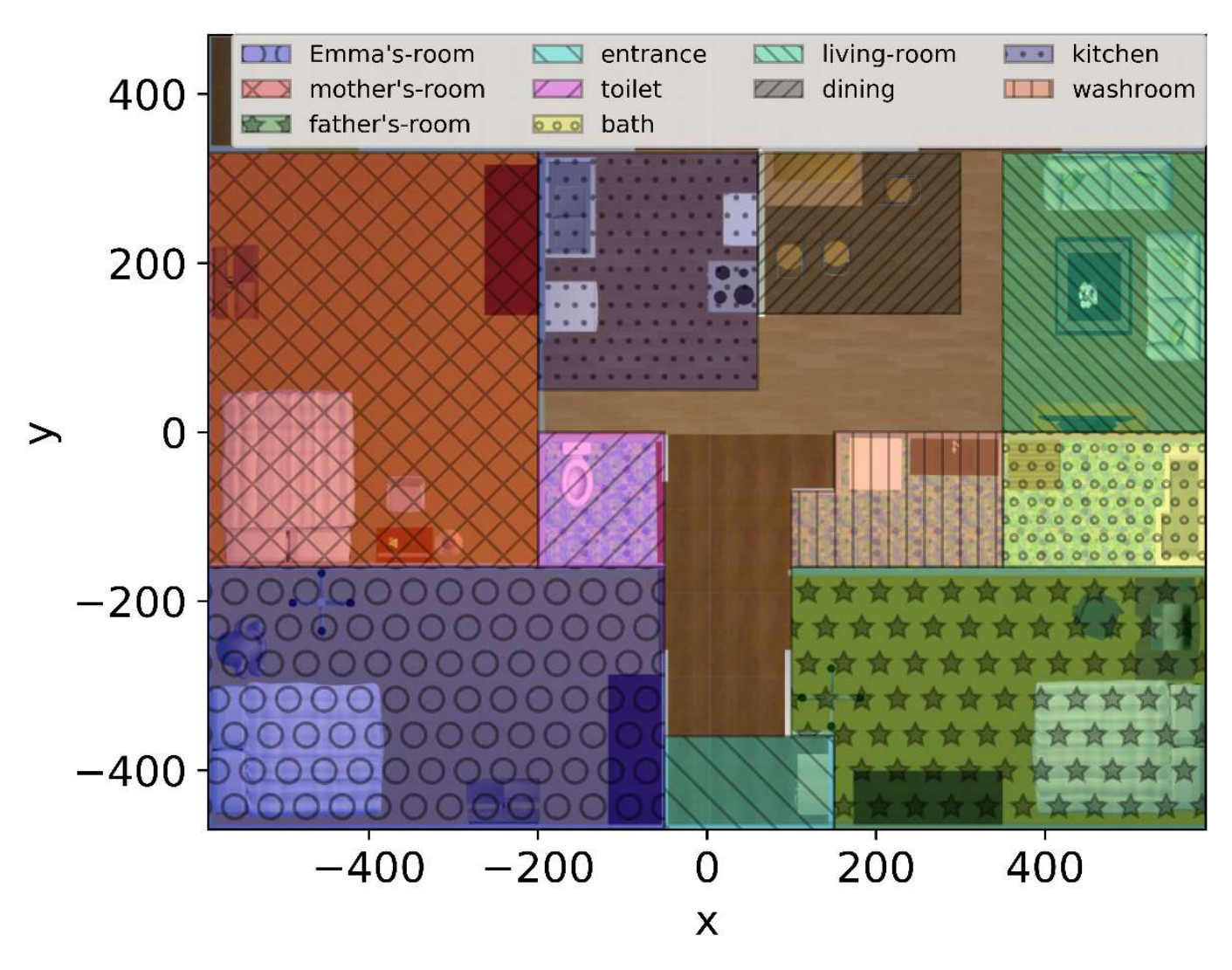}
          \hspace{1.6cm} (a)
        \end{center}
      \end{minipage}
      \begin{minipage}{0.45\hsize}
        \begin{center}
          \includegraphics[clip, width=7.0cm]{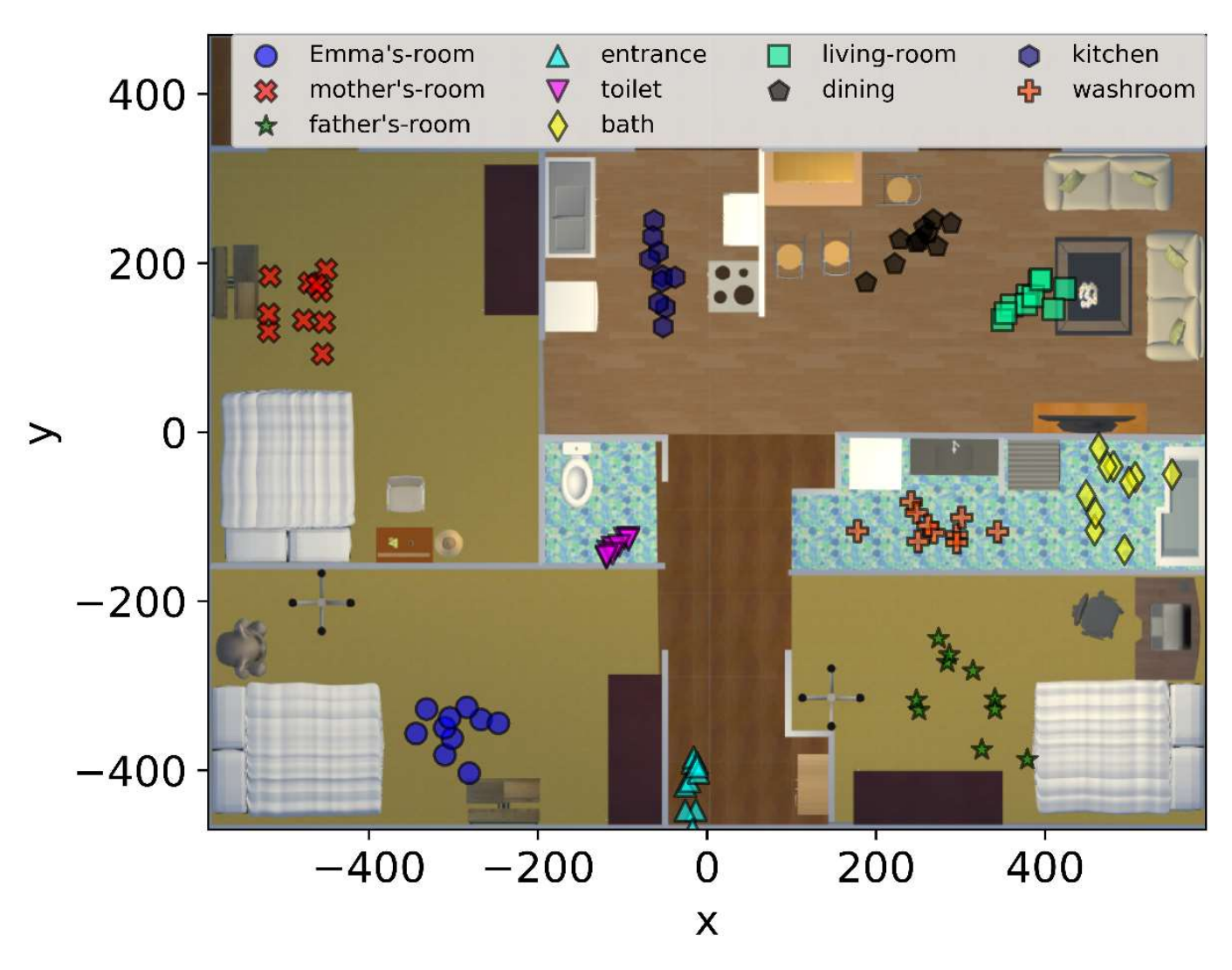}
          \hspace{1.6cm} (b)
        \end{center}
      \end{minipage}
    \end{tabular}
    \caption{(a) Ground truth of regions in a new environment for location names, i.e., Emma's room, mother's room, father's room, entrance, living room, washroom, toilet, dining, bedroom, bath, and kitchen. (b) Results of position prediction for home-specific places in the new environment using the proposed model with sixteen experienced environments.}
    \label{local_position_pre}
  \end{center}
\end{figure}
Fig.~\ref{local_name_pre} shows the experimental result of the qualitative evaluation of adaptive learning of spatial concepts in home-specific places using the proposed model trained with sixteen environments. We confirmed that the proposed model can correctly predict location names for home-specific places such as Emma's room from the test data provided as vision and position information. Furthermore, we confirmed that general places such as kitchens, which are not instructed to the robot in a new environment, can be predicted by the transfer of knowledge using the proposed model. Similarly, the result of position prediction shown in Fig.~\ref{local_position_pre} confirms that all the position data predicted from location names (b) are plotted in regions of ground truth (a).

\begin{figure}[tb]
  \begin{center}
    \includegraphics[width=420pt]{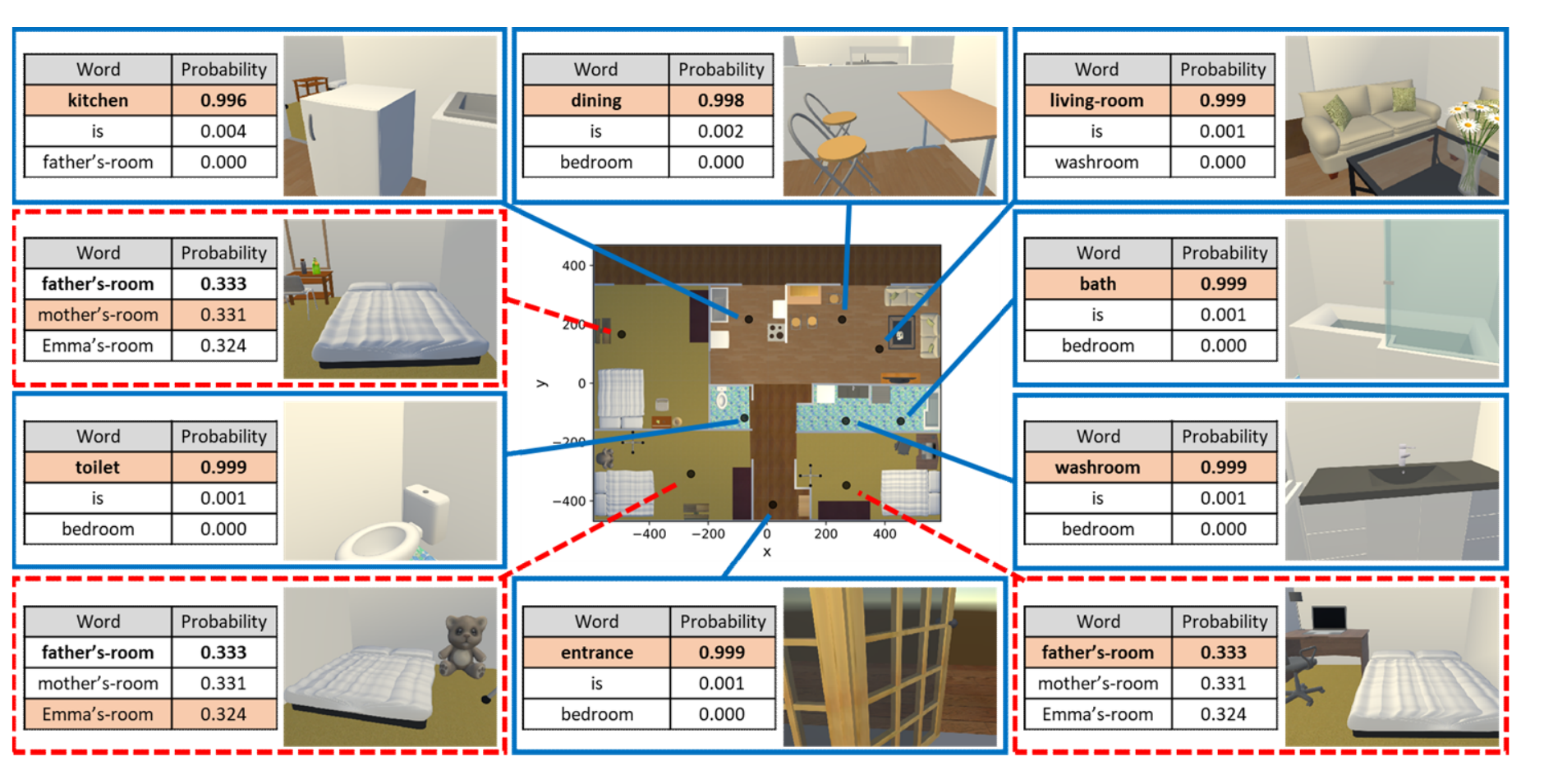}
    \caption{Result of name prediction for home-specific places in a new environment using the proposed model with sixteen experienced environments. The predicted location names are listed as tables of the three best predicted words. The position of the obtained data and the image captured at the position are shown as a black bot on the map and an image in a box, respectively.}
   \label{local_name_pre_miss}
  \end{center}
\end{figure}
However, the predicted names from the given test data for home-specific places were incorrect, in some scenarios (Fig.~\ref{local_name_pre_miss}). The ground truth for the name prediction of the test data indicated by the red dotted line in the upper left of Fig.~\ref{local_name_pre_miss} was mother's room, but the top of the predicted word was father's room. The results of name predictions at the three home-specific places indicated by the red dotted lines show that father's room, mother's room, and Emma's room were predicted with similar probabilities. We considered that this was the result of learning one spatial concept as a category because the features in the images of the three rooms were similar. Such learning results were confirmed in several trials. This was considered to be the reason for the performance of the name and position prediction in the graph of Fig.~\ref{local_position_pre} remaining at 0.5 and 0.6, respectively. In the virtual environment of this experiment, the same furniture was placed in each home-specific place except for one piece of furniture; therefore, the similarity of image features between places was high. Considering this environment, classifying three places into one category as a result of categorization based on multimodal information is natural. In an actual home environment, we predict that the image features will be different from those in the virtual environment used in this experiment; therefore, it is important to evaluate the proposed model in actual environments.

\section{Conclusion}
\label{sec:conclusion}
We propose a hierarchical Bayesian model that transfers the knowledge of spatial concepts formed in experienced environments to the learning of spatial concepts in a new environment. The proposed model enables a robot to predict location names and positions for general places such as kitchens in a new environment by transferring the knowledge from experienced environments to a new environment and to adaptively learn home-specific places, such as `Emma's room,' based on their own observations and linguistic instructions from a user in a new environment.

We evaluated the performance of the proposed model in the generalization of spatial concepts to environments through experiments to learn spatial concepts and transfer knowledge between virtual home environments. The experiment demonstrated that the proposed model can learn about a general place such as a kitchen in a new environment without any instruction from the user and can predict its location name and position with high accuracy. 
In addition, we evaluated the performance of name and position prediction on whether it was possible to adaptively learn both general and home-specific places such as Emma's room based on the robot’s own observations and linguistic instructions from users. The experimental results demonstrated that the proposed model can predict the location name and position by appropriately learning both general and home-specific places.

However, in some scenarios, places such as `Emma's room,' `father's room,' and `mother's room' were learned as one category because of their similar visual features. This can be interpreted as a result of the objects installed in the three rooms of the virtual environment were the same except for one. We are attempting to increase the complexity of virtual environments and perform experiments in diverse and complex actual environments.
In particular, the actual environment contains various furniture and objects, and in such an environment, a framework for integrated learning from image feature extraction to categorization is required. Therefore, we consider using JMVAE~\citep{JMVAE}, which is a multimodal extension of VAE~\citep{VAE}, as an inference algorithm.

In addition to transferring knowledge of location concepts between environments, we are also working on transferring knowledge of spatial concepts between robots using cloud services. SERKET~\citep{SERKET} and NeuroSERKET~\citep{nSERKET} approaches will be used for communication and sharing of model parameters and observations between robots.

\section*{Acknowledgments}
This study was partially supported by the Japan Science and Technology Agency (JST) Core Research for Evolutionary Science and Technology (CREST) research program, under Grant JPMJCR15E3, and by the Japan Society for the Promotion of Science (JSPS) KAKENHI under Grant JP18K18134, and by MEXT Grant-in-Aid for Scientific Research on Innovative Areas 4903 (Co-creative Language Evolution), 17H06383.

\label{lastpage}

\bibliographystyle{tADR}
\bibliography{tADR}

%\begin{comment}

\clearpage
\appendix
\section{Related studies}
\label{apdx:related_studies}
\subsection{Image recognition of places}
In the field of computer vision, several studies on the image recognition of places using CNN models have been performed ~\citep{Zhou14,Zhou18,Ursic16,Qassim18,oyebode19}. 
Zhou et al. proposed CNN models trained using Places, which is a a large-scale dataset containing hundreds of thousands of scene images of indoor and outdoor places~\citep{Zhou14,Zhou18}. 
Qassim et al. proposed Residual Squeeze VGG16~\citep{Qassim18}, a compressed version of the CNN model named VGG16~\citep{VGG16}. The performance of Residual Squeeze VGG16 was demonstrated using a large dataset called Place365-Standard.
Ursic et al. proposed a CNN-based image recognition method for rooms in home environments~\citep{Ursic16}. The place class was estimated by dividing the place images with similar objects and features with the CNN.
Oyebode et al. proposed an environment recognition method~\citep{oyebode19} that combined a CNN and an ontology. The ontology refers to the explicit representation of the upper classes as places and lower classes as objects.
Place-recognition methods that do not use CNNs have also been proposed~\citep{Pronobis06,Kostavelis13,sahdev16}.
Sahdev proposed a place-recognition method based on a histogram of oriented uniform patterns (HOUP) descriptor~\citep{sahdev16}. With this method, place images are classified using principal component analysis and a support vector machine. In the experiment, the place class is estimated in real time from the images acquired by two mobile robots. 

The methods based on image recognition using CNNs and support vector machines have enabled the prediction of a place class from a captured image. However, for a robot to achieve movement based on a linguistic command such as `go to kitchen,' it is required to have a function of associating a place class with a position in the actual world. This study targeted both the prediction of location names and a function to predict positions in the actual world using linguistic information. This function is called semantic mapping.

\subsection{Semantic mapping}
%Survey of semantic mapping
Semantic mapping refers to methods of assigning meanings, such as vocabulary representing places or classes of objects and places, to an environment map held by the robot. A semantic map is a map to which such meanings are assigned. There are various approaches to semantic mapping~\citep{Kostavels15}.
%Semantic mapping based on occupancy grid map
In the early studies on semantic mapping, methods of attaching object labels obtained by deep learning algorithms to an occupancy grid map as semantic attributes were proposed~\citep{Ranganathan07, Rusu09, Rusu08, Espinace13}. These studies enabled a robot to use the semantic information on objects in an occupancy grid map but not to use place information such as location names associated with spatial regions (e.g, kitchens)
Studies on semantic mapping based on deep learning algorithms with labeled datasets for places have been conducted~\citep{Goeddel16, Sunderhauf16, Himstedt17, Brucker18, Robert16, Posada18, Rangel18, Pal19}.
Sunderhauf et al. proposed a place-recognition method by applying a CNN to images obtained by a robot moving in an environment and assigning the obtained place classes to an occupied grid map~\citep{Sunderhauf16}. A large dataset called the Place205 dataset was used to train the CNN.
Pal et al. proposed DEDUCE~\cite{Pal19}, which combined a place-recognition CNN called Place365~\cite{places365} and an object-recognition CNN called You Only Look Once (YOLO)~\cite{yolov3}. Experimental results indicated that this integrated approach provided a higher recognition accuracy than existing place-recognition methods.

%Semantic mapping based on topological map
An approach using a topological map has also been studied for semantic mapping~\cite{Thrun98, Nielsen04, Nozawa13,Rangel17,Rangel18,Rangel19,Balaska20,Hiller19}.
A topological map consists of nodes that store information related to the location, such as vocabulary, images, and positions, and edges that represent transitions between the nodes.
Rangel et al. proposed LexToMap, which uses the CNN class of an object recognizer as the vocabulary label and generates a topological map in which the vocabulary label and the location node are related~\cite{Rangel17,Rangel18,Rangel19}.
Balaska et al. proposed a method of generating semantic maps based on a topological map using an unsupervised learning approach~\cite{Balaska20}. The robot acquires image and position data at various positions as nodes and creates a cluster based on the similarity of image features and the proximity of positions to generate a semantic map. This approach enables the location category to be estimated from the images acquired by the robot when it revisits the learned location.
Hiller et al. proposed a method of dividing space by detecting doors and generating a semantic map~\cite{Hiller19}. A semantic map is created by assigning one class to the space of the occupancy grid map surrounded by the wall and the segmentation mask defined by a door detector based on a CNN. The study also generated a topometric map connecting the center of each class and the door using an edge.

Many of the semantic mapping methods using pre-trained CNN labels can predict general places based on large-scale labeled datasets in advance.
However, applying these methods to home-specific places such as Emma's room is difficult because no labeled dataset exists for home-specific places that have unique location names, features, and positions in each home environment. An approach to generate a semantic map from image information and position information using an unsupervised learning approach has been proposed, but the correspondence between the estimated category and the place-related vocabulary is not learned. For a robot to manage home-specific places with location names and positions, it should have a function to learn places based on natural linguistic information and observations obtained in each environment.
For a robot to learn the location names and positions of home-specific places, this study focused on an approach to learn places from linguistic information and observations obtained in each environment.

\subsection{Robot navigation based on linguistic information}
Vision-and-language navigation (VLN) has been proposed as an approach applying the concept of visual question answering (VQA)~\cite{VQA} to robot navigation.
Studies on VLN achieved room-to-room navigation based on natural linguistic information using a dataset composed of image sequences with 21,576 linguistic instructions for navigation routes~\cite{Anderson18, Krantz20}. 

These studies achieved robot navigation based on linguistic information of routes using a sequence-to-sequence model, which integrates CNN features, long short-term memory (LSTM), and action categories. However, acquiring knowledge of home-specific places through on-site learning in a new environment is difficult because a large-scale dataset is required for image sequences with the linguistic instructions of navigation routes for home-specific places.
For robots to behave adaptively in various home environments, this study aimed to achieve the learning of home-specific places on-site from a small amount of linguistic information using the spatial concept models described below.

\subsection{Spatial concept model}
\label{apdx:SpCoA}
\begin{figure}[tb]
  \begin{center}
    \includegraphics[width=350pt]{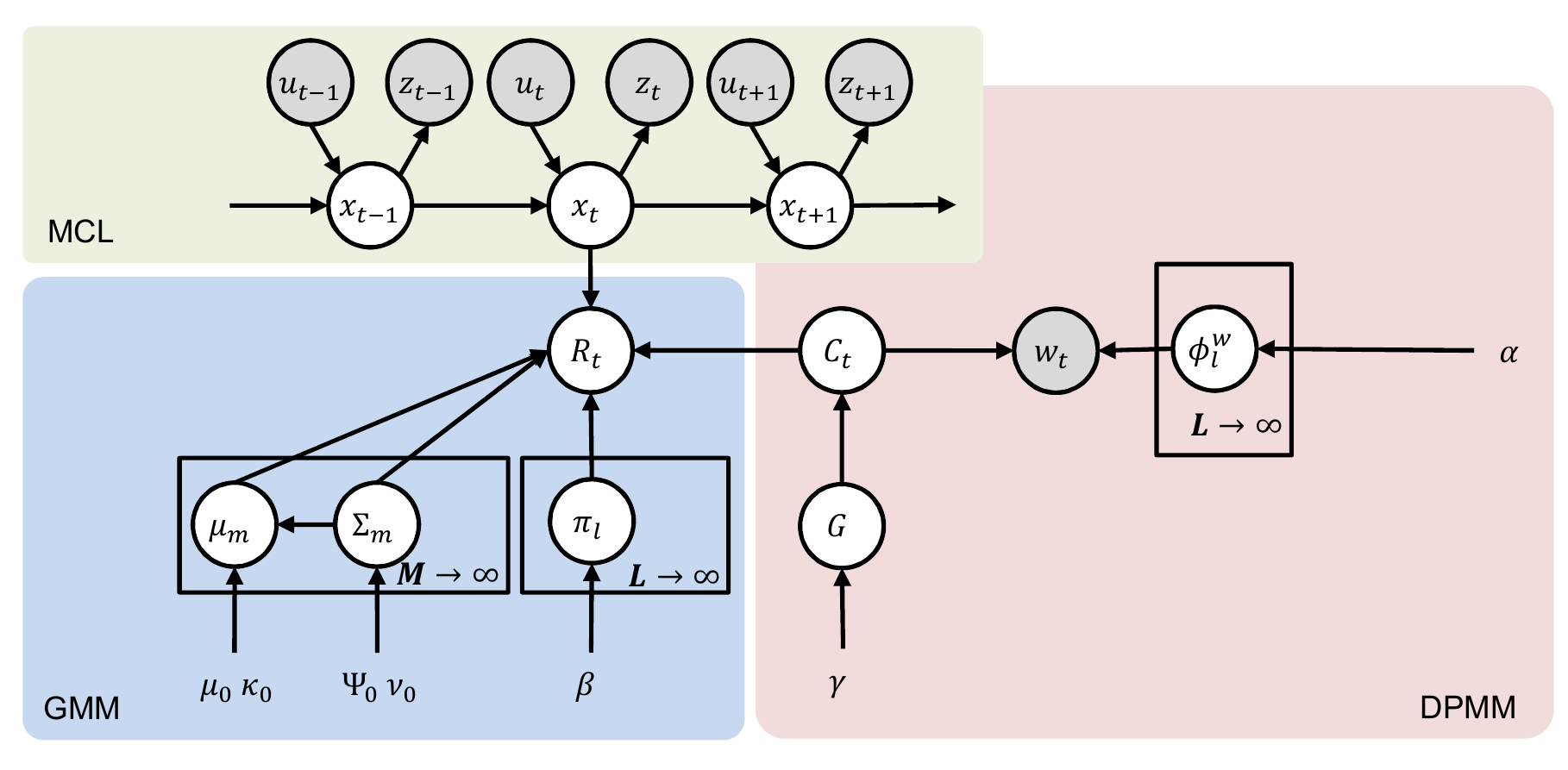}
    \caption{Graphical model of a conventional spatial concept model called SpCoA. SpCoA consists of MCL to estimate a position ($x_t$), Gaussian mixture model (GMM) to infer a spatial region ($R_t$), and a Dirichlet process mixture model (DPMM) to infer a spatial concept $C_t$.}
   \label{conv_model}
  \end{center}
\end{figure}

Spatial concept models have been proposed as place-learning models based on the probabilistic generative process of linguistic instructions from users and observed information obtained in the environment.
The spatial concept is the categorical knowledge of a place formed from multimodal information, e.g., location names, visual features, and spatial areas on a map. 
Taniguchi et al. proposed a nonparametric spatial concept acquisition (SpCoA)~\citep{Taniguchi16} that learns the spatial concept ($C_t$) and spatial region ($R_t$) from the self-position ($x_t$) obtained using Monte Carlo localization (MCL)~\citep{MCL} and a linguistic instruction from the user as a bag of words ($w_t$) (Fig.~\ref{conv_model}).
Fig.~\ref{conv_model} shows the graphical model of SpCoA.
This model has no mechanisms to transfer inferred parameters between environments.
They proposed SpCoNavi~\citep{Taniguchi20}, which estimates the route to the target place based on human linguistic instructions such as `Go to kitchen' by utilizing the spatial concept learned in Online SpCoSLAM~\citep{Taniguchi17}. Isobe et al. proposed a spatial conceptual model that utilized object information as the bag of objects obtained from YOLO9000~\citep{YOLO} instead of the image features~\citep{Isobe17}. This method can learn the relationship between objects and a place and achieve the prediction of the place in which the object should be placed. In the methods introduced above, the spatial region of the place is represented by a Gaussian distribution of positions. Katsumata et al. proposed SpCoMapping, which learns the spatial concept using Markov random fields instead of the Gaussian distribution of positions~\citep{Katsumata19}. With this method, a place can be learned by considering the shape of the environment, such as a wall using the Markov random field. Hagiwara et al. focused on the hierarchy of places and proposed a method of modeling the hierarchical structure of spatial concepts~\citep{Hagiwara18}.

Spatial concept models can learn home-specific places from the linguistic instructions of the user and observations obtained from sensors in the environment. However, even when learning spatial concepts for general places, the user must instruct robot about the places from the beginning, which imposes a heavy load on the user. To reduce the cost of instructions from the user, we consider a method of transferring the knowledge of spatial concepts obtained from other environments to the learning of places in a new environment.

\subsection{Transfer learning in deep learning models}
Transfer learning is an approach that reduces the cost of instruction and learning for machine learning models. Several methods for transfer learning have been proposed for deep learning models.
Deep learning requires many labeled datasets, but collecting labeled datasets is costly. A transfer learning approach has been proposed as a solution to such problems. Transfer learning is a method of transferring knowledge from the source domain to the target domain by relaxing the assumption that the training and test data are independent of each other and follow the same distribution.
Tan et al. classified transfer learning approaches for deep learning into the following four types~\citep{Tan18}.
The first is an instance-based approach that uses weighted instances (observed values, dataset) of the source domain to learn the target domain.
The second is a mapping-based approach that maps the target and source domain instances to a new data space to learn the target domain.
The third is a network-based approach that uses a partial network that has been learned in advance by learning the source domain to learn the target domain.
The fourth is an adversarial-based approach that uses adversarial technology to find transferable features that are suitable for both source and target domains.

The proposed model based on the probabilistic generation process does not correspond to any of the above neural network approaches, but it interprets the dataset obtained in experienced environments as the source domain and the dataset obtained in a new environment as the target domain. This study aimed to transfer knowledge of spatial concepts in a probabilistic generative model.

\subsection{Knowledge transfer in the probabilistic generative model}
Teh et al. proposed the latent Dirichlet allocation using the hierarchical Dirichlet process (HDP-LDA) as models of prior distribution shared between mixture models~\citep{HDP}. Latent Dirichlet allocation (LDA) is a probabilistic generative model for classifying documents using unsupervised learning~\citep{LDA}. HDP-LDA is an extended model that enables a corpus to be shared between documents. HDP-LDA infers the distribution of the topic set of documents by dividing them into training and test documents. This model represents training and testing documents on different plates on a graphical model and achieves the knowledge transfer of the corpus from the domain of the training document set to the domain of the test document set by learning the distribution of the topic set shared between the plates. Experimental results demonstrated that increasing the number of training documents improved the predictive performance of the test documents.
In a study applying the mechanism of knowledge transfer to a spatial concept model, Katsumata et al. proposed spatial concept formation-based semantic mapping with generative adversarial networks (SpCoMapGAN)~\citep{Katsumata20}. This method uses the framework of generative adversarial nets (GANs) to transfer the joint distribution of all cell classes learned from the semantic maps of various known environments to generate semantic maps in a new environment. In this study, the knowledge of room features and positional relationships between rooms was transferred to the generation of semantic maps in a new environment.
As these studies demonstrated, a probabilistic generative model with knowledge transfer learns common knowledge between documents or environments as parameters outside of the plates and achieves high prediction performance in a new document or environment as a target domain by transferring common knowledge.

This study focused on the problem of adaptive learning of home-specific places based on linguistic instructions from users in a new home environment while retaining the knowledge of general places transferred from experienced environments, which was not addressed in a previous study~\citep{Katsumata20}.

\section{Proposed method} 
\label{apdx:proposed_method}
\subsection{Spatial concept formation based on multimodal information}
\label{apdx:proposed_method:multimodal}
\begin{figure}[tb]
\begin{center}
    \includegraphics[width=450pt]{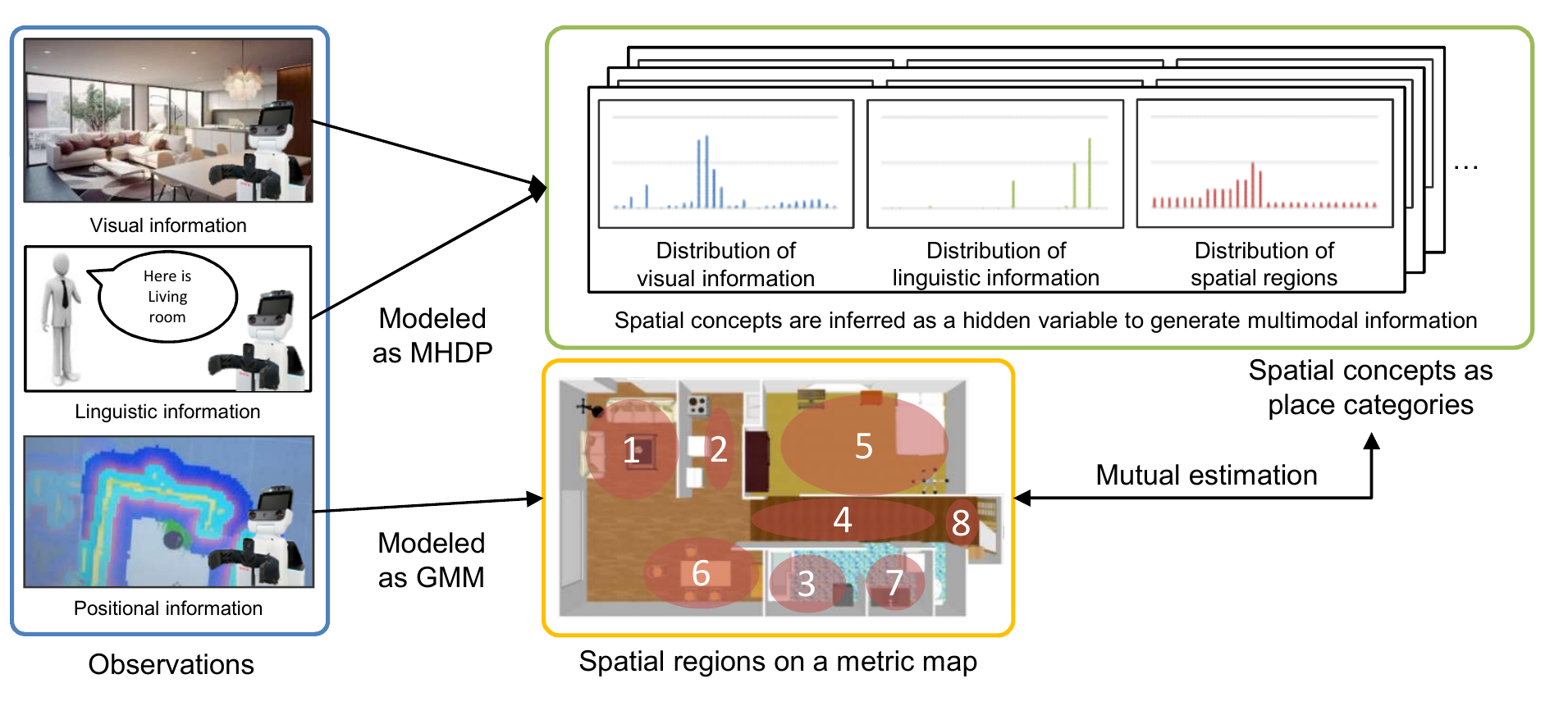}
    \caption{Spatial concept formation based on multimodal information in each environment. The generative process for visual and linguistic information is modeled as the multimodal hierarchical Dirichlet process. The generative process for positional information is modeled as a a Gaussian mixture model (GMM).}
   \label{multimodal}
  \end{center}
\end{figure}
An overview of the spatial concept formation based on multimodal information in each home environment is shown in Fig.~\ref{multimodal}. The proposed model simultaneously estimates spatial and spatial concepts by using positional information estimated by the robot, visual information observed from the robot's camera at the position, and linguistic information provided as the user's linguistic instructions for a place. The linguistic instruction is provided as a sentence involving a location name, such as `Here is living room.' The generative process of spatial regions on a metric map is modeled as a GMM. The generative process of spatial concepts was modeled as an MHDP.

\subsection{Details of the generative process} 
\label{apdx:proposed_method:generate}
The detailed definitions of the generative process of the proposed model are as follows.
\begin{itemize}
\item In Equations \ref{gene1} to \ref{gene3}, $C_{t_e}$ denotes an index of a spatial concept to a data obtained at time $t$ in an environment $e$. $C_{t_e}$ is generated using a categorical distribution with parameter $G_e$. $G_e$ is generated using a Dirichlet process with parameters $G_0$ and $\gamma$ as the stick-breaking process~\cite{HDP}. $G_0$ is generated using a GEM distribution~\citep{GEM} with the parameter $\gamma_0$.
\item In Equations \ref{gene8} to \ref{gene9}, $v_{t_e}$ and $n_{t_e}$ denote visual and linguistic information obtained at time $t$ in an environment $e$, respectively. $v_{t_e}$ and $n_{t_e}$ are defined as vectors using the following equations:
\begin{eqnarray}
v_{t_e} &=& [v_{t_e,1}, v_{t_e,2}, \cdots, v_{t_e,D^v}]\\
w_{t_e} &=& [w_{t_e,1}, w_{t_e,2}, \cdots, w_{t_e,D^w}],
\end{eqnarray}
where $D^v$ and $D^w$ denote the dimensions of $v_{t_e}$ and $w_{t_e}$, respectively.
$v_{t_e}$ and $w_{t_e}$ are generated by a multinomial distribution with the parameters $\theta^v_{C_{t_e}}$ and $\theta^w_{C_{t_e}}$, respectively.

\item In Equations \ref{gene4} to \ref{gene7}, $\phi$ can be described as a generalized parameter to generate observed information for all environments, and $\theta$ can be described as a specific parameter to generate observed information for $e$.
$\theta$ and $\phi$ are defined as follows:
\begin{eqnarray}
\theta^v_{e,l} &=& [\theta^v_{e,l,1},\theta^v_{e,l,2},\cdots,\theta^v_{e,l,D^v}]\\
\theta^w_{e,l} &=& [\theta^w_{e,l,1},\theta^w_{e,l,2},\cdots,\theta^w_{e,l,D^n}]\\
\phi^v_l &=& [\phi^v_{l,1},\phi^v_{l,2},\cdots,\phi^v_{l,D^v}]\\
\phi^w_l &=& [\phi^w_{l,1},\phi^w_{l,2},\cdots,\phi^w_{l,D^n}],
\end{eqnarray}
where $l$ is the number of a spatial concept.

$\mathrm{Dir}(\phi^v_l, \delta^v)$ and $\mathrm{Dir}(\phi^w_l, \delta^w)$ in Equations \ref{gene6} and \ref{gene7} are calculated using the following equations:

\begin{eqnarray}
\mathrm{Dir}(\phi^v_l, \delta^v) &=& \frac{\Gamma{(\sum_{k=1}^{D^v}\delta^v \phi^v_{l,k})}}
{\sum_{k=1}^{D^v} \Gamma{(\delta^v \phi^v_{l,k})}}
\prod_{k=1}^{D^v} (\theta^v_{e,l,k})^{\delta^v \phi^v_{l,k}-1}\\
\mathrm{Dir}(\phi^w_l, \delta^w) &=& \frac{\Gamma{(\sum_{k=1}^{D^w}\delta^w \phi^w_{l,k})}}
{\sum_{k=1}^{D^w} \Gamma{(\delta^w \phi^w_{l,k})}}
\prod_{k=1}^{D^w} (\theta^w_{e,l,k})^{\delta^w \phi^w_{l,k}-1},
\end{eqnarray}
where $\delta^v$ and $\delta^w$ are hyperparameters to control the mixture between $\phi$ as the generalized knowledge in all environments and $\theta$ as the specific knowledge in an environment.

\item In Equations \ref{gene10} to \ref{gene11}, 
the self-position of a robot ($x_{t_e}$) at $t$ in $e$ is defined as follows: 
\begin{eqnarray}
x_{t_e} =[x,y,\sin \theta,\cos \theta],
\end{eqnarray}
where $x,y$, and $\theta$ are coordinate values in the horizontal and vertical directions and an orientation in the coordinate system of a map, respectively.

\item In Equations \ref{gene12} to \ref{gene15}, $\pi_{e,l}$ denotes the parameter of a categorical distribution to generate the index of a spatial region $R_{t_e}$ ($R_{t_e} \in \bm{R_e} =\left\{1,2,3,\cdot \cdot \cdot,M\right\}$) at $t$ in $e$ based on a spatial concept $l$. $\pi_{e,l}$ is defined as follows:
\begin{eqnarray}
\pi_{e,l}&=[\pi_{e,l,1},\pi_{e,l,2},\cdots,\pi_{e,l,M}].
\end{eqnarray}
\item Equation \ref{gene15} is defined as follows:
\begin{equation}
p(R_{t_e} \mid x_{t_e},C_{t_e},\bm{\mu_{e}},\bm{\Sigma_{e}},\bm{\pi_{e}})
= \frac{\mathcal{N}(x_{t_e} \mid \mu_{e,R_{t_e}},\Sigma_{e,R_{t_e}})\mathrm{Cat}(R_{t_e} \mid \pi_{e,C_{t_e}})}{\sum_{R_{t_e}=m} \mathcal{N}(x_{t_e} \mid \mu_{e,m},\Sigma_{e,m})\mathrm{Cat}(m \mid \pi_{e,C_{t_e}})},
\end{equation}
where $\mu_{e,m}$ and $\Sigma_{e,m}$ are the mean vector and covariance matrix at a region $m$, respectively.
\end{itemize}

\subsection{Inference process} 
\label{apdx:proposed_method:gibbs}
In the proposed model, the model parameters are inferred using a Gibbs sampler. The inference process for the parameters is as follows:
\begin{enumerate}
\renewcommand{\labelenumi}{$\langle \arabic{enumi} \rangle$}

\item Initialize latent variables $R_{t_e}$, $C_{t_e}$ with a large integer value.
Initialize the parameters $\theta^v_{e,l}$, $\theta^w_{e,l}$, $G_e$, $\phi^v_l$, and $\phi^w_l$ using a uniform probability.
Initialize the parameters $\pi_{e,l}$ and $G_0$ using the stick-breaking process.
The parameters $\mu_{e,m}$ for the Gaussian distribution in $e$ are selected as the center coordinates of $e$.
Here, $\Sigma_{e,m}$ are initialized to $\Sigma_{e,m} = \rm{diag}[\sigma_{init}, \sigma_{init}, \sigma_{init}, \sigma_{init}]$.

\item Determine the index of $R_{t_e}$ for all data acquired in  $e$.
The index of $R_{t_e}$ is sampled using a posterior distribution as follows:
\label{gibbs1}
\begin{flalign}
R_{t_e} &\sim p(R_{t_e}=m \mid x_{t_e},C_{t_e},\bm{\mu_e},\bm{\Sigma_e},\bm{\pi_e}) \nonumber \\
&\propto p(x_{t_e} \mid \mu_{e,m},\Sigma_{e,m})p(R_{t_e}=m \mid \pi_{e,C_{t_e}}).
\end{flalign}

\item Determine the index of a spatial concept ($C_{t_e}$) for all data acquired in $e$.
The index of $C_{t_e}$ is sampled using a posterior distribution as follows:
\begin{flalign}
C_{t_e} &\sim p(C_{t_e}=l \mid x_{t_e},v_{t_e},w_{t_e},R_{t_e},\bm{\mu_e},\bm{\Sigma_e},\bm{\theta^v_e},\bm{\theta^w_e},\bm{\pi_e},G_e) \nonumber \\
&\propto p(v_{t_e} \mid \theta^v_{e,l})p(w_{t_e} \mid \theta^w_{e,l})p(R_{t_e} \mid \pi_{e,l})p(C_{t_e}=l \mid G_e).
\end{flalign}

\item Estimate the parameters of a Gaussian distribution $\mu_{e,m},\Sigma_{e,m}$ for generating region $m$ in $e$ as follows:

\begin{flalign}
\mu_{e,m}, \Sigma_{e,m} &\sim p(\mu_{e,m},\Sigma_{e,m} \mid R_{t_e}=m,\bm{X_e},\mu_0,\kappa_0,\Psi_0,\nu_0) \nonumber \\
&\propto \mathcal{N}(\bm{X_{e,m}} \mid \mu_{e,m},\Sigma_{e,m})\mathcal{NIW}(\mu_{e,m},\Sigma_{e,m} \mid \mu_0,\kappa_0,\Psi_0,\nu_0) \nonumber \\
&\propto \mathcal{NIW}(\mu_{e,m},\Sigma_{e,m} \mid \mu_{e,m,new},\kappa_{e,m,new},\Psi_{e,m,new},\nu_{e,m,new}),
\end{flalign}
where $\bm{X_e}$ denotes a set of observed position data in $e$, $\bm{X_{e,m}}$ denotes a set of -data assigned to the index of region $m$ in $e$, and $\mathcal{N}(\cdot)$ denotes a Gaussian distribution, $\mathcal{NIW}(\cdot)$ denotes a Gauss-inverse Wishart distribution, and $\mu_{e,m,new},\kappa_{e,m,new},\Psi_{e,m,new},\nu_{e,m,new}$ denote hyperparameters updated based on the conjugation between a Gaussian distribution and a Gauss-inverse Wishart distribution.

\item Estimate the parameter of a multinomial distribution $\theta^v_{e,l}$ for generating visual information ($v_{t_e}$) in environment $e$ based on a spatial concept $l$ as follows:
\begin{flalign}
\theta^v_{e,l} &\sim p(\theta^v_{e,l} \mid \bm{C_e}=l,\bm{V_{e}},\phi^v_l,\delta^v) \nonumber \\
&\propto \mathrm{Mult}(\bm{V_{e,l}} \mid \theta^v_{e,l})\mathrm{Dir}(\theta^v_{e,l} \mid \delta^v\phi^v_l)  \nonumber  \\
&\propto \mathrm{Dir}(\theta^v_{e,l} \mid \phi^v_{e,l,new}),
\end{flalign}
where $\bm{V_e}$ denotes a set of visual data acquired in $e$, $\bm{V_{e,l}}$ denotes a set of visual data allocated to a spatial concept $l$ in $e$, and $\phi^v_{e,l,new}$ denotes a hyperparameter updated based on the conjugation between a multinomial distribution and a Dirichlet distribution.

\item Estimate the parameter of a multinomial distribution $\theta^w_{e,l}$ for generating linguistic information $w_{t_e}$ in an environment $e$ based on $l$ as follows:
\begin{flalign}
\theta^w_{e,l} &\sim p(\theta^w_{e,l} \mid \bm{C_e}=l,\bm{W_{e}},\phi^w_l,\delta^w) \nonumber \\
&\propto \mathrm{Mult}(\bm{W_{e,l}} \mid \theta^w_{e,l})\mathrm{Dir}(\theta^w_{e,l} \mid \delta^w\phi^w_l)  \nonumber  \\
&\propto \mathrm{Dir}(\theta^w_{e,l} \mid \phi^w_{e,l,new}),
\end{flalign}
where $\bm{W_e}$ denotes a set of linguistic data acquired in $e$, $\bm{W_{e,l}}$ denotes a set of linguistic data allocated to $l$ in $e$, and  $\phi^w_{e,l,new}$ denotes a hyperparameter updated based on the conjugation between a multinomial distribution and a Dirichlet distribution.

\item Update the parameter $\pi_{e,l}$ of a multinomial distribution for generating the index of a region based on the index of a concept $l$ in environment $e$ by the following formulas:
\begin{flalign}
\pi_{e,l} &\sim p(\pi_{e,l} \mid \bm{X_e},\bm{C_e}=l,\bm{R_{e}}, \bm{\mu_e},\bm{\Sigma_e},\beta) \nonumber \\
&\propto \mathrm{Cat}(\bm{R_{e,l}} \mid \pi_{e,l})\mathrm{Dir}(\pi_{e,l} \mid \beta) \nonumber  \\
&\propto \mathrm{Dir}(\pi_{e,l} \mid \beta_{e,l,new}),
\end{flalign}
where $\vector{R_e}$ denotes a set of indices of a region in $e$, $\vector{R_{e,l}}$ denotes a set of indices of a region allocated to a spatial concept $l$ in an environment $e$, and $\beta_{e,l,new}$ denotes a hyperparameter updated based on the conjugation between a multinomial distribution and a Dirichlet distribution.

\item Estimate the parameter of the multinomial distribution $G_e$ for generating a spatial concept in the environment $e$ as follows: 
\label{gibbs2}
\begin{flalign}
G_e &\sim p(G_e \mid \bm{C_e},G_0,\gamma) \nonumber \\
&\propto \mathrm{Cat}(\bm{C_e} \mid G_e)\mathrm{Dir}(G_e \mid \gamma G_0)  \nonumber \\
&\propto \mathrm{Dir}(G_e \mid \gamma_{e,new}),
\end{flalign}
where $\gamma_{e,new}$ denotes a hyperparameter updated based on the conjugation between a multinomial distribution and a Dirichlet distribution.

\item Perform steps $\langle$\ref{gibbs1}$\rangle$ to $\langle$\ref{gibbs2}$\rangle$ for each $e$ \label{gibbs3}

\item Estimate the parameter of the multinomial distribution $\phi^v_l$ for generating a parameter $\theta^{v,e}_l$ for $l$ as follows: \begin{flalign}
\phi^v_l &\sim p(\phi^v_l \mid \bm{V},\bm{C_{e}}=l,\alpha^v) \nonumber \\
&\propto \mathrm{Mult}(\bm{V}_l \mid \phi^v_l)\mathrm{Dir}(\phi^v_l \mid \alpha^v)  \nonumber  \\
&\propto \mathrm{Dir}(\phi^v_l \mid \alpha^v_{l,new}),
\end{flalign}
where $\bm{V}$ denotes the set of visual data in all the environments, $\bm{V}_l$ denotes a set of visual data allocated to $l$ in all environments, and $\alpha_{l,new}^v$ denotes a hyperparameter updated based on the conjugation between a multinomial distribution and a Dirichlet distribution.

\item Estimate the parameter of the multinomial distribution $\phi^n_l$ for generating a parameter $\theta^{n,e}_l$ for a spatial concept $l$ as follows:
\begin{flalign}
\phi^w_l &\sim p(\phi^w_l \mid \bm{W},\bm{C_{e}}=l,\alpha^w) \nonumber \\
&\propto \mathrm{Mult}(\bm{W}_l \mid \phi^w_l)\mathrm{Dir}(\phi^w_l \mid \alpha^w)  \nonumber  \\
&\propto \mathrm{Dir}(\phi^w_l \mid \alpha^w_{l,new}),
\end{flalign}
where $\bm{W}$ denotes a set of linguistic data for all the environments, $\bm{W}_l$ denotes a set of linguistic data allocated to the spatial concept $l$ in all environments, and $\alpha_{l,new}^w$ denotes a hyperparameter updated based on the conjugation between a multinomial distribution and a Dirichlet distribution.

\item Update the parameter of a multinomial distribution $G_0$ for generating a parameter $G_e$ as follows: \label{gibbs4}
\begin{flalign}
G_0 &\sim p(G_0 \mid \bm{G},\gamma,\gamma_0) \nonumber \\
& \propto \mathrm{Mult}(\bm{G} \mid \gamma G_0)\mathrm{Dir}(G_0 \mid \gamma_0) \nonumber \\
& \propto \mathrm{Dir}(G_0 \mid \gamma _{0,new}),
\end{flalign}
where $\bm{G}$ denotes a set of all parameters to generate a spatial concept, and $\gamma_{0,new}$ denotes a hyperparameter updated based on the conjugation between a multinomial distribution and a Dirichlet distribution. The weak-limit approximation~\cite{WeakLimit} is used in the inference process.

\item Perform steps $\langle$\ref{gibbs3}$\rangle$ to $\langle$\ref{gibbs4}$\rangle$ for the number of iterations.

\item Construct $\theta^n,\phi^n$ based on the following mutual information.
\begin{flalign}
\nonumber
I(w_{t_e};C_{t_e} \mid \Theta)=&\sum_w \sum_c P(w,c \mid \Theta)\log\frac{P(w,c \mid \Theta)}{P(w \mid \Theta)P(c \mid \Theta)}, \\
=& \sum_w \sum_c P(w \mid \phi^w_{c})P(c \mid G_0)\log\frac{P(w \mid \phi^w_{c})P(c \mid G_0)}{P(c \mid G_0 )\sum_c P(w \mid \phi^w_{c})P(c \mid G_0)}. 
\end{flalign}
When the mutual information $I(w_{t_e};C_{t_e} \mid \Theta)$ between a word $w_{t_e}$ and a category $C_{t_e}$ is smaller than the threshold $\epsilon$ in all categories $C_{t_e}=\{1,2,...,L\}$, 0 is assigned to the probability value of a word $w_{t_e}$ in word distributions $\theta^w$, $\phi^w$. This process is a heuristic process for removing words such as `is' that are not related to places from the location names. In the experiments, the threshold was set as $\epsilon=0.1$.

\end{enumerate} 

\clearpage

\subsection{Algorithm}
\label{apdx:proposed_method:algorithm}
\begin{algorithm}
 \caption{Learning algorithm of the proposed model}
 \begin{algorithmic}[1]
 \STATE Initialize all variables and parameters
 \FOR {$i = 1$ to $iteration~number$}
 \STATE //Gibbs sampling
  \FOR {$e = 1$ to $E$}
   \FOR {$t = 1$ to $T_e$}
    \STATE $R_{t_e} \sim p(R_{t_e}=m \mid x_{t_e},C_{t_e},\bm{\mu_e},\bm{\Sigma_e},\bm{\pi_e})$
    \STATE $C_{t_e} \sim p(C_{t_e}=l \mid x_{t_e},v_{t_e},w_{t_e},R_{t_e},\bm{\mu_e},\bm{\Sigma_e},\bm{\theta^v_e},\bm{\theta^w_e},\bm{\pi_e},G_e)$
   \ENDFOR
   \FOR {$m = 1$ to $M$}
    \STATE $\mu_{e,m},\Sigma_{e,m} \sim p(\mu_{e,m},\Sigma_{e,m} \mid \bm{X_{e}},\bm{C_e},\bm{R_e}=m,\bm{\pi_e},\mu_0,\kappa_0,\Psi_0,\nu_0)$
   \ENDFOR
   \FOR {$l = 1$ to $L$}
    \STATE $\theta^v_{e,l} \sim p(\theta^v_{e,l} \mid \bm{V_{e}},\bm{C_e}=l,\phi^v_l,\delta^v)$
    \STATE $\theta^w_{e,l} \sim p(\theta^w_{e,l} \mid \bm{W_{e}},\bm{C_e}=l,\phi^w_l,\delta^w)$
    \STATE $\pi_{e,l} \sim p(\pi_{e,l} \mid \bm{X_e},\bm{C_e}=l,\bm{R_{e}}, \bm{\mu_e},\bm{\Sigma_e},\beta)$
   \ENDFOR
   \STATE $G_{e} \sim p(G_e \mid \bm{C_e},G_0,\gamma)$
  \ENDFOR
  \FOR {$l = 1$ to $L$}
    \STATE $\phi^v_{l} \sim p(\phi^v_{l} \mid \bm{V},\bm{C}=l,\alpha^v)$
    \STATE $\phi^w_{l} \sim p(\phi^w_{l} \mid \bm{W},\bm{C}=l,\alpha^w)$
   \ENDFOR
   \STATE $G_0 \sim p(G_0 \mid \bm{C},\gamma, \gamma_0)$
   \STATE //Reconstruction $\bm{\theta^w,\phi^w}$ using mutual information
   \FOR {$i = 1 $ to $K$}
   \IF{max~$I(w_{t_e}=D_i;C_{t_e}|\Theta) < \epsilon$}
    \STATE $\{ \{ \theta^w_{e,l,i} \}^E_{e=1} \}^L_{l=1} \leftarrow 0$
    \STATE $\{ \phi^w_{l,i} \}^L_{l=1} \leftarrow 0$
   \ENDIF
  \ENDFOR
  \STATE Normalize $\bm{\theta^w}$,$\bm{\phi^w}$
 \ENDFOR
 \end{algorithmic} 
 \label{algorithm}
\end{algorithm}

\clearpage
\section{SpCoA(+MI)}
\label{apdx:spcoa}
\subsection{Generative process}
\label{apdx:spcoa:generate}
The generative process of the proposed model is described as follows:
The generative process of a category $C_{t}$ is
\begin{flalign}
G  &\sim \mathrm{GEM}(\gamma) \\
C_{t} &\sim \mathrm{Cat}(G)
\end{flalign}
The generative process of observations $w_{t}$ is
\begin{flalign}
\phi^w_l &\sim \mathrm{Dir}(\alpha) \\
w_{t} &\sim \mathrm{Mult}(\phi^w_{C_{t}})
\end{flalign}
The generative process of a region $R_{t}$ is
\begin{flalign}
x_{t} &\sim p(x_{t} \mid x_{t-1},u_{t}) \\
z_{t} &\sim p(z_{t} \mid x_{t}) \\
\pi_{l} &\sim \mathrm{GEM}(\beta) \\
\Sigma_{m} &\sim \mathcal{IW}(\Psi_0,\nu_0) \\
\mu_{m} &\sim \mathcal{N}(\mu_0,\Sigma_{m}/\kappa_0) \\
R_{t} &\sim p(R_{t} \mid x_{t},C_{t},\bm{\mu},\bm{\Sigma},\bm{\pi})\label{formula:spcoa_R}
\end{flalign}
However, Equation \ref{formula:spcoa_R} is defined as follows:
\begin{equation}
p(R_{t} \mid x_{t},C_{t},\bm{\mu},\bm{\Sigma},\bm{\pi})
= \frac{\mathcal{N}(x_{t} \mid \mu_{R_{t}},\Sigma_{R_{t}})\mathrm{Cat}(R_{t} \mid \pi_{C_{t}})}{\sum_{R_{t}=m} \mathcal{N}(x_{t} \mid \mu_{m},\Sigma_{m})\mathrm{Cat}(m \mid \pi_{C_{t}})}. 
\end{equation}

\subsection{Inference process}
\label{apdx:spcoa:gibbs}
In SpCoA, the model parameters are inferred using a Gibbs sampler. The inference process for the parameters is as follows:
\begin{enumerate}
\renewcommand{\labelenumi}{$\langle \arabic{enumi} \rangle$}

\item Initialize latent variables $R_{t}$ and $C_{t}$ with a large integer value.
Initialize parameters $G,\phi^w_l$ using a uniform probability.
Initialize parameters $\pi_{l}$ using a stick-breaking process.
The parameters $\mu_{m}$ for the Gaussian distribution in an environment are selected as the center coordinates of an environment.
Here, $\Sigma_{m}$ is initialized to $\Sigma_{m} = \rm{diag}[\sigma_{init}, \sigma_{init}, \sigma_{init}, \sigma_{init}]$.

\item Determine the index of region $R_{t}$ for all data acquired in environment $e$.
The index of $R_{t}$ is sampled using a posterior distribution as follows:
\label{gibbs3}
\begin{flalign}
R_{t} &\sim p(R_{t}=m \mid x_{t},C_{t},\bm{\mu},\bm{\Sigma},\bm{\pi}) \nonumber \\
&\propto p(x_{t} \mid \mu_{m},\Sigma_{m})p(R_{t}=m \mid \pi_{C_{t}}).
\end{flalign}

\item Determine the index of spatial concept $C_{t}$ for all data acquired in an environment.
The index of $C_{t}$ is sampled using a posterior distribution as follows:
\begin{flalign}
C_{t} &\sim p(C_{t}=l \mid x_{t},w_{t},R_{t},\bm{\mu},\bm{\Sigma},\bm{\phi^w},\bm{\pi},G) \nonumber \\
&\propto p(w_{t} \mid \theta^w_{l})p(R_{t} \mid \pi_{l})p(C_{t}=l \mid G).
\end{flalign}

\item Estimate the parameters of a Gaussian distribution $\mu_{m},\Sigma_{m}$ for generating a region $m$ in an environment as follows:

\begin{flalign}
\mu_{m}, \Sigma_{m} &\sim p(\mu_{m},\Sigma_{m} \mid \bm{X}, R_{t}=m,\mu_0,\kappa_0,\Psi_0,\nu_0) \nonumber \\
&\propto \mathcal{N}(\bm{X_{m}} \mid \mu_{m},\Sigma_{m})\mathcal{NIW}(\mu_{m},\Sigma_{m} \mid \mu_0,\kappa_0,\Psi_0,\nu_0) \nonumber \\
&\propto \mathcal{NIW}(\mu_{m},\Sigma_{m} \mid \mu_{m,new},\kappa_{m,new},\Psi_{m,new},\nu_{m,new}),
\end{flalign}
where $\bm{X}$ denotes a set of observed position data in an environment, $\bm{X_{m}}$ denotes a set of observed position data assigned to the index of region $m$ in an environment, $\mathcal{N}(\cdot)$ denotes a Gaussian distribution, $\mathcal{NIW}(\cdot)$ denotes a Gauss-inverse Wishart distribution, and $\mu_{m,new},\kappa_{m,new},\Psi_{m,new},\nu_{m,new}$ denote hyperparameters updated based on the conjugation between a Gaussian distribution and a Gauss-inverse Wishart distribution.

\item Estimate the parameter of a multinomial distribution $\phi^w_{l}$ for generating linguistic information $w_{t}$ in an environment based on a spatial concept $l$ as follows:
\begin{flalign}
\phi^w_{l} &\sim p(\phi^w_{l} \mid \bm{W},\bm{C}=l,\alpha) \nonumber \\
&\propto \mathrm{Mult}(\bm{W_{l}} \mid \theta^w_{l})\mathrm{Dir}(\theta^w_{l} \mid \alpha)  \nonumber  \\
&\propto \mathrm{Dir}(\theta^w_{l} \mid \alpha_{l,new}),
\end{flalign}
where $\bm{W}$ denotes a set of linguistic data acquired in an environment, $\bm{W_{l}}$ denotes a set of linguistic data allocated to a spatial concept $l$ in an environment, and  $\alpha_{l,new}$ denotes a hyperparameter updated based on the conjugation between a multinomial distribution and a Dirichlet distribution.

\item Update the parameter $\pi_{l}$ of a multinomial distribution for generating the index of a region based on the index of a concept $l$ in an environment by the following formulas:
\begin{flalign}
\pi_{l} &\sim p(\pi_{l} \mid \bm{X},\bm{C}=l,\bm{R}, \bm{\mu},\bm{\Sigma},\beta) \nonumber \\
&\propto \mathrm{Cat}(\bm{R_{l}} \mid \pi_{l})\mathrm{Dir}(\pi_{l} \mid \beta) \nonumber  \\
&\propto \mathrm{Dir}(\pi_{l} \mid \beta_{l,new}),
\end{flalign}
where $\vector{R}$ denotes a set of the indices of a region in an environment, $\vector{R_{l}}$ denotes a set of indices of a region allocated to a spatial concept $l$ in an environment, and $\beta_{l,new}$ denotes a hyperparameter updated based on the conjugation between a multinomial distribution and a Dirichlet distribution.

\item Estimate the parameter of the multinomial distribution $G$ for generating a spatial concept in the environment as follows: 
\label{gibbs4}
\begin{flalign}
G &\sim p(G \mid \bm{C},\gamma) \nonumber \\
&\propto \mathrm{Cat}(\bm{C} \mid G)\mathrm{Dir}(G \mid \gamma)  \nonumber \\
&\propto \mathrm{Dir}(G \mid \gamma_{new}),
\end{flalign}
where $\gamma_{new}$ denotes a hyperparameter updated based on the conjugation between a multinomial distribution and a Dirichlet distribution.

\item Perform steps from $\langle$\ref{gibbs3}$\rangle$ to $\langle$\ref{gibbs4}$\rangle$ 

\item Construct $\phi^w$ using the following mutual information.
\begin{flalign}
\nonumber
I(w_{t};C_{t} \mid \Theta)=&\sum_w \sum_c P(w,c \mid \Theta)\log\frac{P(w,c \mid \Theta)}{P(w \mid \Theta)P(c \mid \Theta)}, \\
=& \sum_w \sum_c P(w \mid \phi^w_{c})P(c \mid G)\log\frac{P(w \mid \phi^w_{c})P(c \mid G)}{P(c \mid G)\sum_c P(w \mid \phi^w_{c})P(c \mid G)}. 
\end{flalign}
When the mutual information $I(w_{t};C_{t} \mid \Theta)$ between a word $w_{t}$ and a category $C_{t}$ is smaller than the threshold $\epsilon$ in all categories $C_{t}=\{1,2,...,L\}$, 0 is assigned to the probability value of a word $w_{t}$ in a word distribution $\phi^w$. This process is a heuristic process for removing words such as `is' that are not related to places from the location names. In the experiments, the threshold was set as $\epsilon=0.1$.

\end{enumerate} 

\subsection{Algorithm}
\label{apdx:spcoa:algorithm}
\begin{algorithm}
 \caption{Learning algorithm of SpCoA(+MI)}
 \begin{algorithmic}[1]
 \STATE Initialize all variables and parameters
 \FOR {$i = 1$ to $iteration~number$}
 \STATE //Gibbs sampling
   \FOR {$t = 1$ to $T$}
    \STATE $R_{t} \sim p(R_{t}=m \mid x_{t},C_{t},\bm{\mu},\bm{\Sigma},\bm{\pi})$
    \STATE $C_{t} \sim p(C_{t}=l \mid x_{t},w_{t},R_{t},\bm{\mu},\bm{\Sigma},\bm{\phi^w},\bm{\pi},G)$
   \ENDFOR
   \FOR {$m = 1$ to $M$}
    \STATE $\mu_{e,m},\Sigma_{e,m} \sim p(\mu_{e,m},\Sigma_{e,m} \mid \bm{X},\bm{C},\bm{R}=m,\bm{\pi},\mu_0,\kappa_0,\Psi_0,\nu_0)$
   \ENDFOR
  \FOR {$l = 1$ to $L$}
    \STATE $\phi^w_{l} \sim p(\phi^w_{l} \mid \bm{W},\bm{C}=l,\alpha)$
    \STATE $\pi_{l} \sim p(\pi_{l} \mid \bm{X},\bm{C}=l,\bm{R}, \bm{\mu},\bm{\Sigma},\beta)$
   \ENDFOR
   \STATE $G \sim p(G \mid \bm{C},\gamma)$
   \STATE //Reconstruction $\bm{\phi^w}$ using mutual information (Only SpCoA+MI)
   \FOR {$i = 1 $ to $K$}
   \IF{max~$I(w_{t}=D_i;C_{t}|\Theta) < \epsilon$}
    \STATE $\{ \phi^w_{l,i} \}^L_{l=1} \leftarrow 0$
   \ENDIF
  \ENDFOR
  \STATE Normalize $\bm{\phi^w}$
 \ENDFOR
 \end{algorithmic} 
 \label{algorithm_spcoa}
\end{algorithm}

\clearpage
\section{Conditions of experiment for generalization}
\label{apdx:experiment_g}
\subsection{Collection process of multimodal information data}
\label{apdx:experiment_g:data}
Multimodal information as the observations for computational models is collected by the following process.
\begin{enumerate}
\renewcommand{\labelenumi}{$\langle \arabic{enumi} \rangle$}
  \item The user instructs a location name using a sentence at an arbitrary position in a virtual home environment.
  \item The current position on the map is calculated based on Eq. (\ref{eq:position}) and memorized as the positional information ($x_t$).
  \item The robot captures an RGB image using a camera installed on its head and memorizes the visual information ($v_t$) calculated using Eq. (\ref{eq:vision}) using the captured images.
  \item A linguistic instruction from the user as a sentence is converted into a bag-of-words expression using Eq. (\ref{eq:word}) and is memorized as the linguistic information ($w_t$). In the experiment, linguistic instructions for places were provided as text input instead of voice input. Linguistic instructions from the user were provided using different types of sentences. Table~\ref{tab:sentences} shows an example of the sentences provided as linguistic instructions.
  \item The information obtained in steps 2, 3, and 4 is used as a set of multimodal information for observations at time step $t$ in the computational models.
  \item The user operates the robot for a next position to instruct.
\end{enumerate}

\subsection{Details of comparison models}
\label{apdx:experiment_g:models}
\subsubsection{SpCoA}
The spatial concept acquisition (SpCoA) model is an original computational model for acquiring spatial concepts based on multimodal information. SpCoA includes a language model that estimates words from phonemes, but we used a trained language model to unify the experimental conditions with the proposed model. We used SpCoA as the baseline for the proposed model. SpCoA was trained using image and position information in new environments.
$\alpha_n=0.1$, $\gamma=10.0$, $\beta=1.0$, $\kappa_0=0.05$, $\Psi_0=\mathrm{diag}[10.0,10.0,0.5,0.5]$, $\nu_0=10.0$, $\mu_0$ was set to the center of a metric map. 

\subsubsection{SpCoA+MI}
Since the proposed model includes the calculation of mutual information in Eq. (\ref{eq:MI}),  SpCoA with mutual information (SpCoA+MI), including the calculation of mutual information, was also prepared as a baseline.
SpCoA+MI was trained with image and position information in new environments.
$\alpha_n=0.1$, $\gamma=10.0$, $\beta=1.0$, $\kappa_0=0.05$, $\Psi_0=\mathrm{diag}[10.0,10.0,0.5,0.5]$, $\nu_0=10.0$, $\mu_0$ was set to the center of a metric map. 
The generative process, inference process, and algorithm of SpCoA+MI are described in Appendix~\ref{apdx:spcoa}.

\subsubsection{Proposed model with 0, 1, 2, 4, 8, and 16 experienced environments}\label{sec:Param}
To evaluate the change in the performance of name prediction and position prediction with the number of experienced environments used for knowledge transfer, we prepared the proposed model trained in 0, 1, 2, 4, 8, and 16 experienced environments. The proposed models were trained with image and position information in new environments and multimodal information in experienced environments.
The Hyperparameters of the proposed model were set as follows: 
$\alpha^v=3.0$, $ \alpha^n=1.0 \times 10^{-2}$, $\delta^v=3.0 \times 10^5$, $\delta^n=1.0 \times 10^4$, $\gamma=10$, $\gamma_0=0.2$, $\beta=3.0$, $\kappa_0=5.0 \times 10^{-2}$, $\Psi_0=\mathrm{diag}[10.0,10.0,0.5,0.5]$, $\nu_0=10$, $\mu_0$ was set to the center of a metric map. 

\subsection{Details of the inference of model parameters from the dataset}
\label{apdx:experiment_g:inference}
In the proposed model, zero to sixteen environments were used for experienced environments and were randomly selected from nineteen experienced environments. One validation environment was used to set the hyperparameters of the comparison models. In experienced environments, all data include position and visual information with language information as linguistic instruction from the user. In the new environment, because the robot must predict location names from position and visual information or to predict positions from the location name with linguistic instructions from the user, the position and visual information were only provided to the robot as observations.

In the comparison models, spatial concepts and position distributions were inferred using Gibbs sampling based on a weak-limit approximation~\cite{WeakLimit} of the stick-breaking process (SBP)~\cite{SBP}, which is a constitutive method of the Dirichlet process ~\citep{HDP}. The upper limits of the spatial concepts and position distributions were set to $L=15$ and $K=20$, respectively. The number of iterations for Gibbs sampling was set to $200$. 

\subsection{Details of evaluation criteria}
\label{apdx:experiment_g:criteria}
In the name prediction, the accuracy of location names $A_{n}$ predicted from visual information and position information was evaluated using the following equation:
\begin{flalign}
A_{n} = \frac{\sum_{l=1}^L \frac{c_l}{N_l}}{L},
\label{eq:name_acc}
\end{flalign}

where $l$ is the index of a place, $L$ is the number of places in the test dataset of a new environment, $c_l$ is the number of correct data in location names predicted by the model, and $N_l$ is the number of test data in a place $l$. $N_l$ was set to 20 in the experiment.
Location names in new environments provided by the user were used as the ground truth of the test data. 

In position prediction, the accuracy of positions $A_{p}$ predicted from word information was evaluated using the following equation:
\begin{flalign}
A_{p} = \frac{\sum_{l=1}^L \frac{c_l}{P_l}}{L},
\label{eq:pos_acc}
\end{flalign}

where $l$ is the index of a place, $L$ is the number of places in the test dataset, $P_l$ is the number of predicted data in each place $l$, and $c_l$ is the number of correct data in location names predicted by the model. $P_l$ was set to 10 in the experiment.

%\end{comment}

\end{document}